\documentclass[a4paper,fleqn]{cas-sc}

\usepackage[numbers, sort&compress]{natbib}
\usepackage{amsmath,amssymb,amsfonts}
\usepackage{algorithmic}
\usepackage{graphicx}
\usepackage{textcomp}
\usepackage{subcaption}
\usepackage{multirow}
\usepackage{multicol}
\usepackage{tablefootnote}
\usepackage{tabularx}
\usepackage{threeparttable}
\usepackage{booktabs}
\usepackage{bm}
\def\tsc#1{\csdef{#1}{\textsc{\lowercase{#1}}\xspace}}
\tsc{WGM}
\tsc{QE}
\tsc{EP}
\tsc{PMS}
\tsc{BEC}
\tsc{DE}

\begin{document}
\let\WriteBookmarks\relax
\def\floatpagepagefraction{1}
\def\textpagefraction{.001}
\shorttitle{MESTI-MEGANet}
\shortauthors{Luu Tu et~al.}

\title [mode = title]{Apex-Centered Spatio-Temporal Rank Pooling and Gradient Attention for Micro-Expression Recognition}                      
\tnotemark[1,2]



\author[1]{Luu Tu Nguyen}
\credit{Conceptualization of this study, Methodology, Software}

\author[1]{Vu Tram Anh Khuong}
\credit{Conceptualization of this study, Methodology, Software}

\author[1]{Thanh Ha Le}
\credit{Conceptualization of this study, Methodology, Software}

\author[1]{Thi Duyen Ngo}
\cormark[1]
\ead{duyennt@vnu.edu.vn }
\credit{Conceptualization of this study, Methodology, Software}








\affiliation[1]{organization={Faculty of Information Technology, VNU University of Engineering and Technology},
                addressline={144 Xuan Thuy, Cau Giay, Ha Noi}, 
                city={Hanoi},
                postcode={100000}, 
                country={Viet Nam}}
\cortext[cor1]{Corresponding author}

  
\begin{abstract}
Micro-expression recognition (MER) is a challenging task due to the subtle and fleeting nature of micro-expressions. Traditional input modalities, such as Apex Frame, Optical Flow, and Dynamic Image, often fail to adequately capture these brief facial movements, resulting in suboptimal performance. In this study, we introduce the Micro-expression Spatio-Temporal Image (MESTI), a micro-expression-specific reformulation of dynamic rank pooling that transforms a video sequence into a single image while emphasizing the onset-apex-offset temporal pattern of micro-expressions. Additionally, we present the Micro-expression Gradient Attention Network (MEGANet), which incorporates a proposed Gradient Attention block to enhance the extraction of fine-grained motion features from micro-expressions. By combining MESTI and MEGANet, we aim to establish a more effective approach to MER. Extensive experiments were conducted to evaluate the effectiveness of MESTI, comparing it with existing input modalities across regular architectures. Moreover, we demonstrate that replacing the input of previously published MER networks with MESTI leads to consistent performance improvements. The performance of MEGANet is also evaluated, showing that our proposed network achieves state-of-the-art results on the SMIC-HS, SAMM and competitive performance on CASMEII datasets, it also achieves leading performance in the reported cross-dataset evaluation settings. The combination of MESTI and MEGANet consistently outperforms the compared methods. These findings underscore the potential of MESTI as a superior input modality and MEGANet as an advanced recognition network, aiming to more effective MER systems in a variety of applications.
\end{abstract}




\maketitle

\section{INTRODUCTION}

\label{sec:introduction}
Facial expression, a vital channel of non-verbal communication, encompasses two primary types: macro expressions and micro expressions. Macro expressions are typically deliberate, easily observable, and last for an extended period, conveying a person's emotions openly \cite{Ekman01051992}. In contrast, micro-expressions (MEs) are brief, involuntary facial movements that last less than 0.5 seconds \cite{Matsumoto2011,Yan2013}, making them significantly challenging to control or fabricate. Unlike macro expressions, MEs reveal a person’s genuine emotions, often surfacing when one attempts to conceal their true feelings \cite{Ekman2003-eb}. These fleeting expressions are especially revealing in high-risk situations \cite{Goh2020, 7851001}, where concealing emotions is common. 
Since these unique characteristics of MEs, they have garnered significant attention as a channel for uncovering individuals' genuine thoughts and emotions. Their involuntary nature provides valuable insights, making them highly applicable in a range of critical fields. For instance, MEs play a crucial role in enhancing the accuracy of deception detection systems, providing valuable insights that more prolonged expressions may not capture \cite{Yildirim2023}. In criminal investigations, law enforcement officers can assess a suspect’s truthfulness by analyzing MEs that may contradict verbal statements \cite{Frank2015}. Beyond security applications, MEs are increasingly relevant in healthcare, particularly in clinical settings, where they can provide essential clues about a patient's emotional state and aid medical professionals in assessing recovery progress \cite{Endres2009}.

Despite its potential, ME recognition (MER) presents significant challenges due to the brevity and subtle intensity of MEs. Studies have shown that even experts achieve only 47\% accuracy in recognizing MEs, highlighting the inherent complexity of this task \cite{frank2009see}. However, leveraging advancements in computational capabilities, as well as modern machine learning and deep learning algorithms, computer-based systems for ME analysis have demonstrated significant superiority over human performance, with accuracy rates often exceeding 50\%. These advancements offer a promising pathway for achieving more accurate and reliable recognition of MEs across a wide range of applications \cite{9915437}.


Early MER approaches primarily relied on handcrafted descriptors, such as texture-based \cite{6130343}, and optical-flow based methods \cite{7286757, happy2017fuzzy, LIONG201882}, to encode subtle facial motion. While these methods established important baselines, their representation capacity is limited and their performance often degrades under complex real-world conditions or in the cross database evaluation scenarios. With the success of deep learning, recent MER research has increasingly focused on learning-based approaches \cite{9915437}. Existing deep MER methods can be broadly understood from two complementary perspectives: input representation and recognition architecture \cite{9915437}. On the one hand, the input representation determines how the weak temporal dynamics of a micro-expression are exposed to the network. On the other hand, the recognition architecture determines whether the model can capture and amplify the subtle motion cues embedded in that representation.


However, current input modalities for MER still suffer from some limitations. Apex-frame-based methods \cite{8451376, 10.1109/FG.2019.8756544} are compact and efficient, but they ignore temporal evolution and therefore miss the onset-to-offset dynamics of micro-expressions. Optical-flow based methods \cite{gan2019off, happy2017fuzzy} provide explicit motion information, but they are often sensitive to noise and unstable when motion is subtle. Dynamic imaging based methods \cite{bilen2016dynamic, 9194324, ACImage}, which compress a sequence into a single image, are attractive because they provide compact spatio-temporal encoding, yet most existing formulations are inherited from general action recognition and are not explicitly tailored to the characteristic onset-apex-offset structure of micro-expressions. Consequently, they may fail to sufficiently emphasize the short discriminative phase around the apex.

At the architectural level, many recent MER networks have improved performance by using motion magnification \cite{Kumar2019ClassificationOF}, graph reasoning \cite{lo2020mer}, or transformer-style context modeling \cite{9747232}. Although these advances are important, two issues remain insufficiently addressed. First, many methods still rely on input representations that are not specifically designed for the temporal morphology of micro-expressions. Second, existing attention mechanisms often emphasize semantic or regional importance at higher feature levels, but do not explicitly target the weak local intensity transitions that are crucial for MER at an early representation stage.

Motivated by these observations, we propose a new MER framework consisting of two complementary components: Micro-expression Spatio-Temporal Image (MESTI) and Micro-expression Gradient Attention Network (MEGANet). MESTI is a compact video-to-image representation designed specifically for MER. Instead of using a generic temporal ranking scheme, it introduces an apex-centered temporal encoding strategy that reflects the onset-apex-offset temporal modeling of micro-expressions, thereby concentrating representational emphasis around the most informative motion phase. MEGANet is a recognition network designed to work effectively with such subtle motion representations. It incorporates a Gradient Attention Block to highlight weak local intensity transitions and Residual Attention Blocks to further model spatial dependencies and refine discriminative facial patterns.

The proposed framework is motivated by an important principle for MER, an effective representation should not only compress the sequence, but should do so in a way that respects the unique temporal structure of micro-expressions, likewise, an effective network should not only model context, but should explicitly focus subtle motion cues that are easy to overlook. 

The main contributions of this paper are summarized as follows:
\begin{itemize}
    \item We propose MESTI, a MER-oriented spatio-temporal representation that reformulates sequence-to-image encoding using an apex-centered ranking principle aligned with the temporal dynamics of micro-expressions.
    \item We propose MEGANet: By integrating a novel Gradient Attention Block and Residual Attention Block, we develop a ME network capable of focusing on motion regions, thereby improving the performance of ME recognition.
    \item A comprehensive set of experimental scenarios is designed to validate the effectiveness of the proposed components, achieving performance that outperforms previous state-of-the-art studies.
\end{itemize}
Through extensive experiments, the effectiveness of each proposed component (MESTI, MEGANet) is demonstrated by evaluating their individual contributions and their combination with previously published methods. The results show that each component of our proposed method enhances the performance of the ME recognition process, and when combined, MESTI and MEGANet yield a effective overall MER approach.

\section{RELATED WORKS}
\subsection{Handcrafted Methods for Micro-Expression Recognition}
Early MER research was dominated by handcrafted feature design. A major line of work focused on texture-based descriptors, where local texture changes over space and time were encoded using methods such as LBP-TOP \cite{6130343} and its variants \cite{inter, 10.1007/978-3-319-16865-4_34}. These approaches were motivated by the observation that subtle facial movements can be reflected as local spatio-temporal texture variations. Subsequent extensions improved the descriptive ability of such features by incorporating different local operators, quantized local patterns \cite{HUANG2016564}. These methods played an important role in the early development of MER because they offered interpretable motion descriptors and established initial benchmarks on public datasets.

Another major line of handcrafted MER methods relied on optical flow and related motion descriptors \cite{7286757, of01, LIONG201882}. Since micro-expressions are fundamentally defined by subtle facial motion, optical flow provides a natural mechanism for capturing local direction and magnitude changes between frames. Methods based on main directional motion statistics, weighted optical flow, and fuzzy directional histograms have shown that motion-oriented handcrafted features are often more suitable for MER than purely static facial appearance features.

Despite their historical importance, handcrafted methods have intrinsic limitations. Their feature extraction process is manually designed and therefore has limited adaptability to the diversity and complexity of spontaneous micro-expressions. In addition, texture descriptors may fail to capture highly localized motion phases, while optical-flow based descriptors are often sensitive to noise, head movement, and illumination changes. As a result, their performance still limited, especially when compared with more recent deep learning approaches.

\subsection{Input Representations in Deep MER}
\subsubsection{Apex frame and static image based representations}
A straightforward way to simplify MER is to use the apex frame. Apex-based representations are computationally efficient and reduce the sequence modeling problem to a standard image classification setting. Several MER methods have demonstrated that the apex frame alone can contain useful discriminative information, especially when combined with powerful feature extractors or local-global fusion mechanisms \cite{8451376, li2020joint}.

However, the major limitation of apex-based input is that it removes most of the temporal evolution of the expression. Micro-expressions are not purely static events; they unfold through a short onset–apex–offset process. When only a single frame is used, the model loses the dynamic context needed to distinguish subtle motion patterns, especially when different classes exhibit similar apex appearances. Thus, apex-frame-based methods are compact, but they sacrifice temporal fidelity.

\subsubsection{Optical flow based deep representations}
To preserve motion information more explicitly, many deep MER methods adopt optical flow or onset-apex motion maps as input. These approaches have proven effective because they directly encode motion magnitude and direction, often providing stronger cues than raw appearance images. Representative methods such as OFF-ApexNet \cite{gan2019off} and subsequent optical flow based networks showed that motion-driven inputs can significantly improve MER performance \cite{Zeng2023, happy2017fuzzy}.

Nevertheless, optical flow is not an optimal solution for MER. Because the movements involved in micro-expressions are extremely weak, the estimated flow field can be unstable and noisy. Small facial movements may be easily contaminated by illumination variation, compression artifacts, or slight head motion. In addition, multi-stream processing of horizontal and vertical flow components often increases model complexity. Therefore, although optical flow introduces temporal information, it may also introduce noise and computational overhead.

\subsubsection{Dynamic imaging based representations}
An appealing alternative is to summarize the entire video sequence into a single spatio-temporal image. Dynamic Image is a representative technique in this category and was originally developed for action recognition \cite{bilen2016dynamic}. It encodes temporal evolution through rank pooling and produces a compact image-like representation that can be processed by conventional CNNs. This idea is attractive for MER because it balances compactness and temporal encoding.

Several MER studies have adopted or adapted this idea, leading to variants such as Affective Image \cite{9194324} and Active Image \cite{ACImage}. These methods demonstrate that sequence-to-image representations can be effective for MER, especially when paired with dedicated CNN architectures. However, most of these methods are inherited from generic dynamic summarization frameworks and are not explicitly designed around the characteristic temporal morphology of micro-expressions. In particular, micro-expressions are not simply temporally progressive events; their discriminative information is concentrated around a short apex-centered phase. A generic temporal compression strategy may therefore dilute or misrepresent the brief motion pattern that is most informative for MER.

This limitation motivates the need for a sequence-to-image representation that is specifically aligned with the onset–apex–offset structure of micro-expressions rather than with generic temporal ordering alone.

\subsection{Deep MER Architectures}
\subsubsection{CNN-based and attention-based networks}
With the rise of deep learning, CNN-based architectures became the dominant paradigm in MER. Early CNN-based methods mainly focused on extracting discriminative appearance features from apex frames or dynamic representations \cite{liu2020offset}. Later studies incorporated attention mechanisms to guide the model toward more informative facial regions. For example, micro-attention \cite{wang2020micro} and magnification-adaptive networks \cite{9747232} attempted to improve MER by focusing on subtle motion-relevant areas. Other methods, such as LEARNet \cite{8844867} and CMNet \cite{wei2023cmnet}, explored more specialized architectures for dynamic or contrastive MER learning.

These methods have significantly advanced the field. However, many of them place attention at intermediate or high-level feature stages and are primarily designed to improve region selection or context modeling. They do not necessarily provide an explicit mechanism for enhancing the weak local intensity transitions that characterize micro-expressions at the representation level. As a result, very subtle motion cues may still be underrepresented during early feature extraction.
\subsubsection{Transformer and hybrid methods}
More recently, MER has benefited from transformer-based and hybrid CNN-transformer models. Methods such as Micron-BERT \cite{Nguyen2023MicronBERTBF} and other recent vision-transformer-based frameworks show that long-range spatial or temporal dependency modeling can be beneficial for MER \cite{PAN2023106258}. Multi-scale attention frameworks and hybrid feature-fusion models have also demonstrated strong performance by combining local motion descriptors with global contextual modeling \cite{HE2025128372}.

\subsection{Research Gaps}
These methods confirm that richer dependency modeling is valuable. However, they still face the core challenge of MER: the signal itself is weak. Even a powerful global modeling framework may not fully solve the problem if the input representation does not expose the subtle motion clearly enough or if the network lacks an explicit mechanism to emphasize weak local transitions at an early stage.

The above review indicates that MER research has made substantial progress, yet two important gaps remain. First, input representation is still a bottleneck. Apex-based inputs are compact but discard temporal evolution; optical-flow-based inputs encode motion but are often noise-sensitive; and existing dynamic-imaging approaches provide compact sequence summarization but are mostly inherited from generic temporal pooling frameworks without explicitly modeling the onset–apex–offset dynamics of micro-expressions.

Second, network design is still not fully aligned with the weak-signal nature of MER. Existing architectures improve contextual reasoning through attention, graph modeling, magnification, or transformers, but many do not explicitly emphasize subtle local intensity transitions in a representation-aware manner.

These two gaps motivate the proposed framework in this paper. MESTI is designed to provide a compact spatio-temporal representation that is explicitly centered on micro-expression dynamics, while MEGANet is designed to amplify weak gradient-based motion cues and refine them through residual attention modeling. Together, they aim to address both the representation-level and architecture-level limitations of existing MER methods.

\section{PROPOSED METHOD}
\begin{figure}
    \centering
    \includegraphics[width=0.8\linewidth]{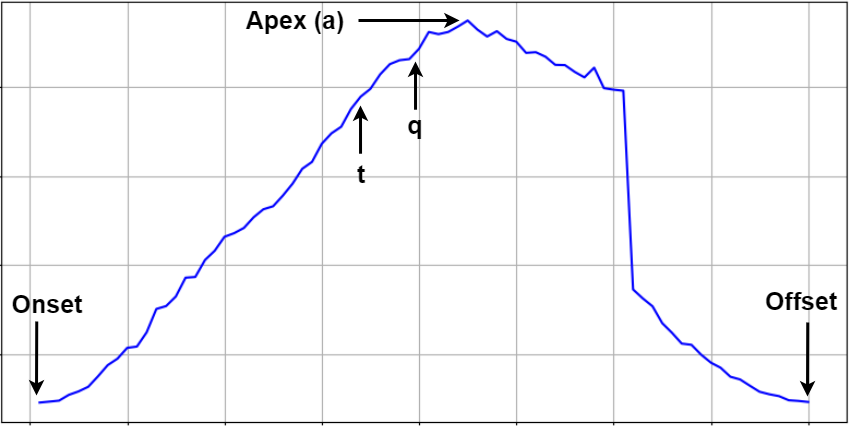}
    \caption{Motion intensity in ME.}
    \label{fig:motion}
\end{figure}
\subsection{Micro-expression Spatio-Temporal Image}
The initial idea for creating an effective input representation for ME stemmed from observing and studying the motion characteristics of MEs. The intensity of motion gradually increases from the onset frame (the starting frame) to the apex (the frame with the highest ME intensity), then decreases towards the offset frame (the final frame representing the ME). Based on this characteristic, the proposed method simulates this motion in the process of constructing a distinctive representation for MEs, namely MESTI. Our objective is to create a spatio-temporal image that effectively represents a ME video. To achieve this, a temporal encoding approach introduced that transforms the entire video sequence into a single representative image. Additionally, our method incorporates the process of aggregating spatial information from the video into a compact static representation.

Inspired by the approximate rank pooling method, which has been used in modeling video evolution \cite{Fernando_2015_CVPR}, a similar strategy is proposed to encode the temporal evolution of MEs into a single image. This approach captures the dynamic variations in facial expressions over time while preserving the spatial structure necessary for effective ME recognition.

Different from conventional Dynamic Image construction, which is based on a temporally monotonic ranking principle, our objective is to reformulate the ranking relation according to the temporal morphology of micro-expressions. A typical micro-expression is not a uniformly progressive event; instead, its discriminative motion is concentrated around a short apex-centered interval, with intensity increasing from onset to apex and decreasing from apex to offset. Therefore, rather than assigning importance purely according to temporal order, MESTI ranks frames according to their proximity to the apex. This change is central: it shifts the inductive bias of rank pooling from generic temporal progression to apex-centered motion concentration, which is more suitable for MER.

\subsubsection{\textbf{Spatial Encoding}}
A video is represented as a sequence of consecutive frames, denoted as $I_1,\dots ,I_t, \dots, I_T$, where $T$ is the total number of frames, and $I_t$ represents the frame at time step $t$. Let $\psi(I_t) \in \mathbb{R}^d$ denote the feature vector extracted from each individual frame $I_t$. In this study, $\psi(I_t)$ is a vector that directly encodes the RGB components of each pixel in the frame $I_t$.

Let $d \in \mathbb{R}^d$ be defined as a parameter vector responsible for assigning a score to each frame \((S(t|d))\) at time $t$ using a ranking function in Equation \ref{eq0}.
\begin{equation} \label{eq0}
    S(t|d) = \langle d, \psi(I_t) \rangle
\end{equation}

The parameter $d$ is learned based on the entire frame sequence, ensuring that the scores assigned to each frame reflect their relative ranking. The learning process of $d$ is formulated as a convex optimization problem using RankSVM\cite{smola2004tutorial}, \(d^*\) refers to the optimal parameter vector \(d\) that is learned based on the entire frame sequence, as described in Equation \ref{eq1}.
\begin{equation} \label{eq1}
d^* = \rho(I_1, \dots, I_T; \psi) = \arg \min_d E(d)
\end{equation}
 
This process integrates spatial information from individual frames into a ME image that preserves structural and appearance details. By leveraging the extracted RGB feature vectors, the method ensures that spatial characteristics of each frame are considered in the ranking process, allowing the network to learn an optimal frame-ordering that reflects their relative importance in the sequence.

\subsubsection{\textbf{Temporal Encoding}}
Temporal encoding is performed based on the characteristic motion patterns of MEs, which serve as a basis for assigning scores to each frame during the rank pooling process of spatial encoding. Figure \ref{fig:motion} illustrates the intensity of motion in ME. The motion characteristics of MEs can be easily observed: the intensity gradually increases from the first frame (onset), peaks at the apex frame, and then gradually decreases toward the final frame (offset) of the ME. Therefore, in this study, we aim to model the motion characteristics of MEs within the temporal encoding process to construct a ME image from the video.

Temporal encoding is implemented by generating a ranking score that simulates the motion intensity of the ME in a straightforward manner during the rank pooling process. Let $I_a$ defined as the apex frame, where the motion intensity of the ME reaches its maximum. Given any two frames $I_{q}, I_{t}$, the frame closer to the apex frame is assigned a higher ranking score in our ranking function.

Thus, for any pair of frames $\{I_{q}, I_{t}\}$ such that $|a-q| \leq |a-t|$, establishing the ranking score: \(S(q|d) > S(t|d)\). Accordingly, Equation \ref{eq1} is further expanded as Equation \ref{eq2}:
\begin{equation} \label{eq2}
\begin{aligned}
    E(d) &= \frac{\lambda}{2} \|d\|^2 + \\
         &\frac{2}{T(T-1)} \times \sum_{|a-q| \leq |a-t|} \max\{0, 1 - S(q|d) + S(t|d)\}.
\end{aligned}
\end{equation}
\begin{figure*}
    \centering
    \includegraphics[width=1\linewidth]{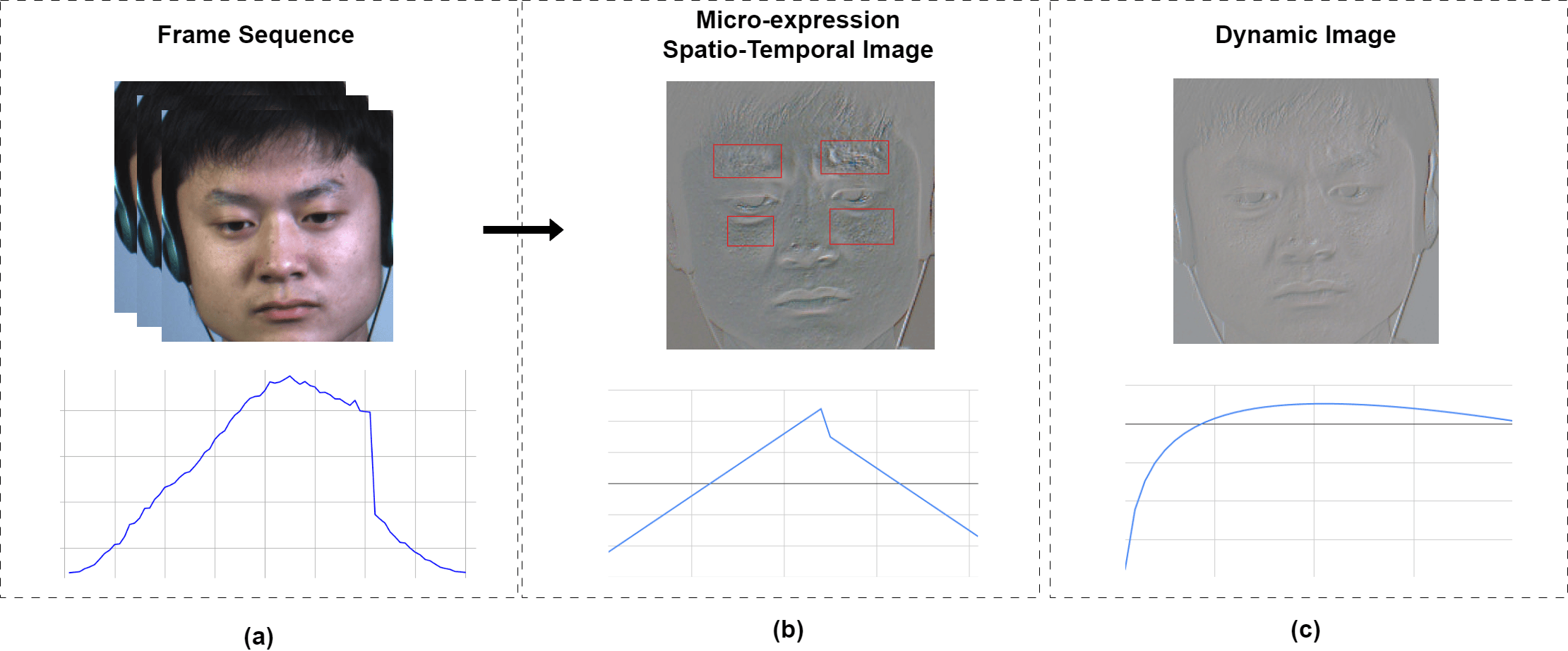}
    \caption{Comparison between conventional Dynamic Image and the proposed MESTI in terms of temporal weighting. Unlike Dynamic Image, which follows a monotonic temporal ranking, MESTI produces an apex-centered coefficient profile that reflects the onset–apex–offset evolution of micro-expressions.}
    \label{fig:enter-label}
\end{figure*}

The first term in Equation \ref{eq2} is the standard quadratic regularizer used in SVMs. The second term is a hinge-loss function that soft-counts how many pairs of frames are incorrectly ranked by the scoring function. To solve the equations involving Equation 1 and Equation 2, the ARP method \cite{bilen2016dynamic} is used. Starting with $d = \Vec{0}$, the first approximated solution obtained by gradient descent is:
\[
d^* = \Vec{0} - \eta \nabla E(d) |_{d = \mathbf{\Vec{0}}} \propto - \nabla E(d) |_{d = \mathbf{\Vec{0}}} \text{ for any } \eta > 0
\]
where:
\[
\nabla E(\Vec{0}) \propto \sum_{|a-q| > |a-t|} \nabla \max\{0, 1 - S(q|d) + S(t|d)\} |_{d = \mathbf{\Vec{0}}}\]
\[
= \sum_{|a-q| > |a-t|} \nabla \langle d, \psi(I_{t}) - \psi(I_{q}) \rangle 
= \sum_{|a-q| > |a-t|} (\psi(I_{t}) - \psi(I_{q})) 
\]

$d^*$ can be expanded as follows:
\[
d^* \propto \sum_{|a-q| > |a-t|} (\psi(I_{q}) - \psi(I_{t}))\] 
\[
= 
\begin{cases}
\sum_{q > t} (\psi(I_{q}) - \psi(I_{t})) & \text{if }  1 \leq t \leq a \\
\sum_{q > t} (\psi(I_{t}) - \psi(I_{q})) & \text{if }  a < t \leq T 
\end{cases}
\]
\[
= 
\begin{cases}
\sum_{t = 1}^{a} \alpha_{t}\psi(I_{t}) & \text{if }  1 \leq t \leq a \\
\sum_{t = a+1}^{T} \alpha_{t}\psi(I_{t}) & \text{if }  a < t \leq T
\end{cases}
\]
where $\alpha_{t}$ is scalar coefficients. By expanding the sum:\\
\textit{When the action in the range of onset frame and apex frame (\(1 \leq t \leq a\)):}

\(\sum_{q > t} \psi(I_{q}) - \psi(I_{t}) = (\psi(I_{2}) - \psi(I_{1}))\)

\[
+ (\psi(I_{3}) - \psi(I_{1})) + (\psi(I_{3}) - \psi(I_{2}))
\]
\[
+ \dots +
\]
\(
(\psi(I_{a}) - \psi(I_{1})) + (\psi(I_{a}) - \psi(I_{2})) + ... + (\psi(I_{a}) - \psi(I_{a-1}))
\)

\textit{When the action in the range of apex frame and offset frame (\(a < t \leq T\)):}

\(\sum_{q > t} \psi(I_{t}) - \psi(I_{q}) = (\psi(I_{a+1}) - \psi(I_{a+2}))\)

\[
+ (\psi(I_{a+1}) - \psi(I_{a+3})) + (\psi(I_{a+2}) - \psi(I_{a+3}))
\]
\[
+ \dots +
\]
\(
(\psi(I_{a+1}) - \psi(I_{T})) + (\psi(I_{a+2}) - \psi(I_{T})) + ... + (\psi(I_{T-1}) - \psi(I_{T}))
\)

Finally, the coefficient \(\alpha_{t}\) can be efficiently computed in two scenarios by aggregating the coefficients of \(\psi(I_{t})\) along with their respective positive and negative signs:
\[
\alpha_t = \begin{cases}
(t-1) - (a - t) & \text{ if }  1 \leq t \leq a \\
(T - t) - (t - a - 1) & \text{ if }  a < t \leq T 
\end{cases}
\]
\begin{equation}\label{eqalpha}
\Rightarrow{}
\alpha_t = \begin{cases}
2t - a - 1 & \text{ if }  1 \leq t \leq a \\
T - 2t + a + 1 & \text{ if }  a < t \leq T 
\end{cases}
\end{equation}

Hence \(d^*\) can be present as the rank pooling operator after using ARP calculation: 
\begin{equation}
    d^* \approx \hat{\rho}(I_1, \dots, I_T; \psi) = \begin{cases}
\sum_{t = 1}^{a} \alpha_{t}\psi(I_{t}) & \text{if }  1 \leq t \leq a \\
\sum_{t = a+1}^{T} \alpha_{t}\psi(I_{t}) & \text{if }  a < t \leq T
\end{cases}
\end{equation}

Finally, the MESTI construction is approximated by multiplying the feature vector representing the RGB component of each frame at time \(t\) with the \(\alpha\) coefficient provided in Equation \ref{eqalpha}. 

The MESTI construction result is shown in Figure 2b using frame sequence (Figure 2a) as input. From the input frame sequence, we represent and observe the motion intensity of the ME representation and have the graph below. The MESTI construction results show that, firstly, our method generates a ranking function that better simulates the nature of the ME motion. Second, through the visual representation results, MESTI has shown more clearly the action units in the ME on the final image constructed compared to the traditional dynamic image method as shown in Figure 2c.

\subsubsection{\textbf{Theoretical properties of the proposed apex-centered ranking}}

\textit{\textbf{Proposition 1. Under the apex-centered pairwise ranking relation}}

$$|a - q| \leq |a - t| \Rightarrow S(q|d) > S(t|d),$$
the first-order approximate rank pooling solution can be written as
$$d^* \propto \sum_{t=1}^{T} \alpha_t \psi(I_t),$$where$$\alpha_t = \begin{cases} 2t - a - 1, & 1 \leq t \leq a, \\ T - 2t + a + 1, & a < t \leq T. \end{cases}$$
Therefore, the coefficient sequence is analytically induced by the proposed ranking relation rather than manually selected.\\
\textit{\textbf{Proof.}}

Starting from the objective in Eq. (3), the first-order approximation of rank pooling gives$$d^* \propto \sum_{|a-q| > |a-t|} (\psi(I_q) - \psi(I_t)).$$By grouping terms with respect to each $\psi(I_t)$, the coefficients for frames in the onset-to-apex interval and the apex-to-offset interval can be collected separately. The resulting piecewise linear form is exactly Eq. (4). Hence, the coefficients are a direct consequence of the apex-centered ranking formulation.\\\\
\textit{\textbf{Proposition 2.}} The coefficient sequence $\{\alpha_t\}_{t=1}^T$ induced by MESTI is apex-centered and unimodal: it increases monotonically on $[1, a]$ and decreases monotonically on $(a, T]$.\\
\textit{\textbf{Proof.}} 

For $1 \leq t < a$:$$\alpha_{t+1} - \alpha_t = 2 > 0,$$which shows monotonic increase toward the apex. 

For $a < t < T$:$$\alpha_{t+1} - \alpha_t = -2 < 0,$$which shows monotonic decrease after the apex. Therefore, the coefficient profile is unimodal and centered around the apex. 

Proposition 2 clarifies why the proposed formulation differs from conventional Dynamic Image ranking. Instead of following global temporal order, MESTI concentrates representational emphasis around the apex, where discriminative ME motion is expected to be strongest. This property makes the representation better aligned with the onset–apex–offset nature of micro-expressions.
\begin{figure*}
    \centering
    \includegraphics[width=1\linewidth]{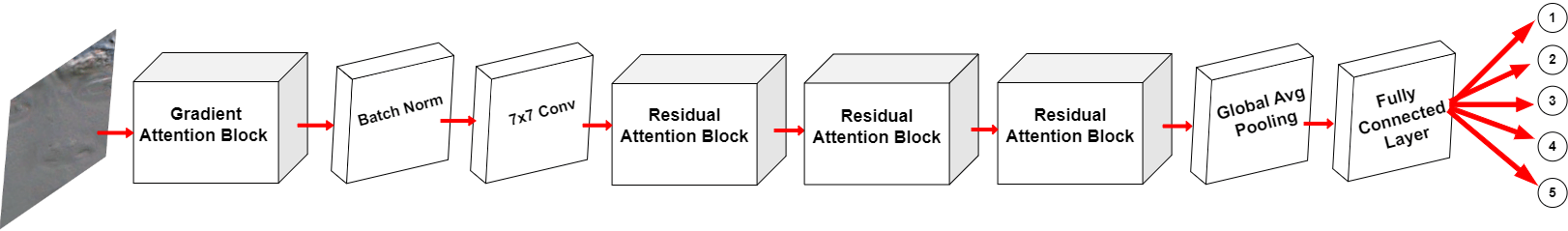}
    \caption{The architecture of the proposed ME Gradient Attention Network (MEGANet)}
    \label{fig:mega}
\end{figure*}

\subsection{Micro-expression Gradient Attention Networks}
The challenge in ME recognition lies in capturing the subtle, transient spatiotemporal patterns that characterize MEs, which often involve subtle intensity changes that conventional CNNs struggle to detect. These expressions are fleeting, making it difficult for traditional methods to effectively focus on the most critical regions of motion. To address this, MEGANet is proposed, a MER network that aims to enhance the detection of MEs by directing attention to areas with significant gradient changes. The core idea behind MEGANet is to combine gradient-guided attention with spatial self-attention, enabling the network to focus on both fine-grained motion transitions and the broader spatial context.

The proposed architecture consists of two key components as shown in Figure 3:  the Gradient Attention Block and the Residual Attention Block. The Gradient Attention Block focuses on amplifying micro-intensity transitions by computing both horizontal and vertical gradients to identify regions with sharp intensity changes. This block generates an attention map through convolution and sigmoid activation, which is then multiplied with the input, enabling the network to prioritize areas with significant micro-movement. The Residual Attention Block, on the other hand, further refines the features by considering the spatial context, ensuring that important structural information is preserved during the feature extraction process. The overall network follows a structured pipeline comprising multiple processing layers:

\begin{itemize}
\item \text{Input layer:} The input to the network is an RGB image of size $224 \times 224 \times 3$.

\item \text{Gradient Attention Block:} Computes spatial gradients to enhance subtle ME features. A convolutional layer followed by a sigmoid activation generates an attention map, which is multiplied with the original input to highlight key regions.  

\item \text{Convolutional Feature Extraction:} A $7 \times 7$ convolutional layer with 64 filters, followed by batch normalization, ReLU activation, and max pooling, extracts low-level spatial features from the input image.  

\item \text{Residual-Attention Blocks:} Three residual attention blocks process the feature maps hierarchically. Each block consists of two $3 \times 3$ convolutional layers, batch normalization, ReLU activation, and a residual connection. A self-attention module is integrated to capture long-range spatial dependencies.  

\item \text{Global Feature Aggregation:} A global average pooling layer compresses the spatial feature maps into a compact feature vector, significantly reducing the number of parameters while retaining crucial information.  

\item \text{Fully Connected Layer and Classification:} The final feature vector is passed through a fully connected (FC) layer and a softmax activation function.  
\end{itemize}

This architecture effectively captures ME dynamics by leveraging gradient-based attention and residual learning, improving the network’s ability to recognize subtle facial movements.

\subsubsection{Gradient Attention Block}
The motivation of gradient-based attention in MEGANet is closely related to the visual structure produced by MESTI. By construction, MESTI transforms a micro-expression video into a compact image in which subtle temporal motion is encoded as localized intensity transitions, especially around motion-relevant facial regions near the eyebrows, eyes, nose, and mouth. Therefore, the discriminative signal in MESTI is not only contained in global appearance, but also in weak local contrast changes and deformation boundaries. Since image gradients respond directly to such local intensity transitions, gradient-based attention is a natural choice for exploiting the motion-sensitive patterns exposed by MESTI. In this sense, the Gradient Attention Block serves as a representation-aware mechanism: it does not simply apply generic attention to the input, but specifically amplifies the subtle motion traces that MESTI is designed to highlight. This design makes MESTI and the Gradient Attention Block complementary: MESTI exposes subtle motion in image form, while gradient attention selectively enhances the most motion-informative regions within that representation.

\begin{figure}
    \centering
    \includegraphics[width=0.9\linewidth]{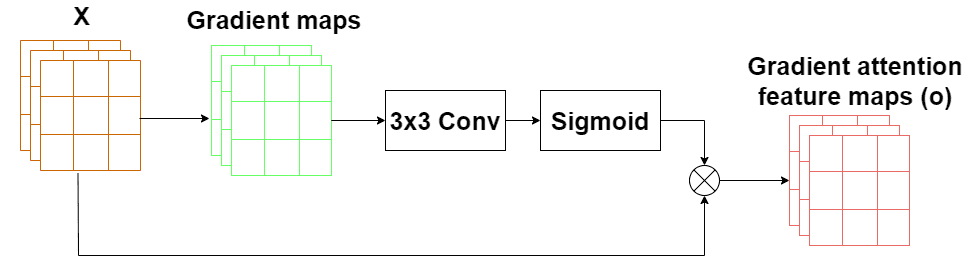}
    \caption{Gradient Attention Block}
    \label{fig:enter-label}
\end{figure}
This block, illustrated in Figure 4,  explicitly models horizontal and vertical intensity gradients to localize ME regions. Given an input image \(X \in R^{B \times C\times H\times W}\), horizontal and vertical gradients at the spatial location \((i,j)\) are computed as: 
\begin{equation} \label{eq3}
\begin{aligned}
    G_x(i,j) &= \|X(i,j+1) - X(i,j)\|_{\text{pad}}\\
    G_y(i,j) &= \|X(i+1;j) - X(i,j)\|_{\text{pad}}\\
\end{aligned}
\end{equation}
where \(G_x, G_y \in R^{B \times C\times H\times W} \) and \(\|\cdot\|_{pad}\) denotes zero-padded absolute differences.
Combined gradient maps are generated through element-wise summation:
\begin{equation} \label{eq4}
\begin{aligned}
    G_{\text{combined}} &= G_x \oplus G_y
\end{aligned}
\end{equation}
The gradient map is then processed through a learnable 3x3 convolutional filter \(W_g\) (\( W_g \in R^{1 \times C\times 3\times 3} \)), followed by sigmoid activation:
\begin{equation} \label{eq5}
    F_{attn} =\sigma(W_g*G_{combined})
\end{equation}
The final output is obtained via element-wise multiplication:
\begin{equation} \label{eq6}
    Y = X \odot F_{attn} 
\end{equation}
This attention map emphasizes regions with significant intensity transitions critical for ME analysis. Figure 5 illustrate the gradient attention map constructed from our proposed MESTI as input image and gradient attention block.
\begin{figure}
    \centering
    \includegraphics[width=0.8\linewidth]{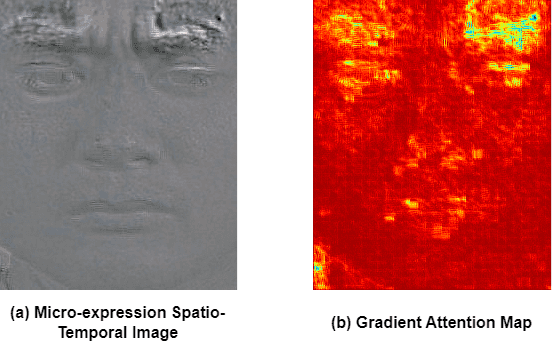}
    \caption{The corresponding Gradient Attention map is generated with the input MESTI.}
    \label{fig:enter-label}
\end{figure}

\subsubsection{Residual - Attention Block}
\begin{figure}
    \centering
    \includegraphics[width=0.65\linewidth]{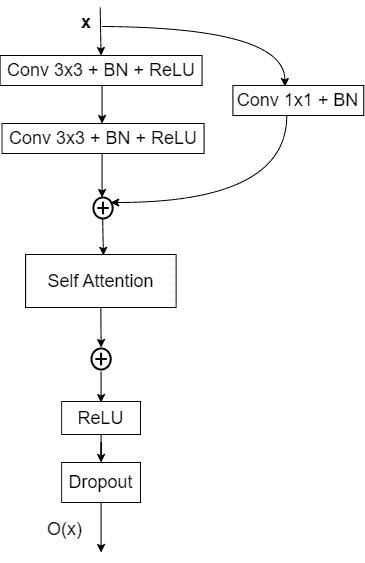}
    \caption{Residual-Attention Block}
    \label{fig:enter-label}
\end{figure}
\begin{figure}
    \centering
    \includegraphics[width=0.9\linewidth]{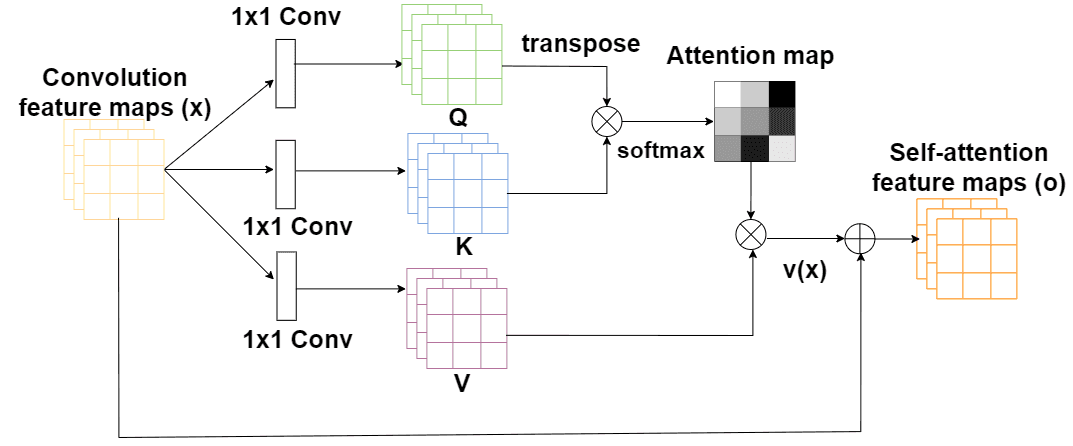}
    \caption{Self-attention based on SAGAN}
    \label{fig:enter-label}
\end{figure}
Our Residual - Attention Block is illustrated in Figure 6, building upon residual connection and SAGAN's self-attention \cite{zhang2019selfattentiongenerativeadversarialnetworks}, this block aims to integrate self-attention into a residual framework to enhance spatial context modeling.
Let \(F(X)\) denote the transformation by two convolutional layers:
\begin{equation} \label{eq7}
\begin{aligned}
    F(X) = BN_2(Conv_2(RELU(BN_1(Conv_1(X)))))\\
    Conv_1: R^{B\times C_{in} \times H \times W} \xrightarrow{} R^{B\times C_{out} \times H' \times W'} \\
     Conv_2: R^{B\times C_{out} \times H' \times W'} \xrightarrow{} R^{B\times C_{out} \times H' \times W'} \\
\end{aligned}
\end{equation}
A shortcut connection handles dimension mismatches:
\begin{equation}
    X_{shortcut} = \left\{\begin{matrix}
Conv_{1 \times 1}(X) \\
X 
\end{matrix}\right.
\end{equation}
The residual output becomes:
\begin{equation}
    X_{res} = X_{shortcut} + F(X)
\end{equation}

Followed by Self-Attention Module proposed by SAGAN illustrated in Figure 7, specifically:
\begin{equation}
\begin{aligned}
    Q &= Conv_{1 \times 1}(X_{res}), \quad Q \in \mathbb{R}^{B \times \frac{C}{8} \times HW}  \\
    K &= Conv_{1 \times 1}(X_{res}), \quad K \in \mathbb{R}^{B \times \frac{C}{8} \times HW} \\
    \mathcal{E} &= \text{softmax}(Q^T K)\\
    V &= Conv_{1 \times 1}(X_{res}), \quad V \in \mathbb{R}^{B \times C \times HW} \\
    Y_{attn} &= \gamma (V\mathcal{E}^T), \quad  \gamma \text{\ is \ learnable}
\end{aligned}
\end{equation}
Finally: 
\begin{equation}
    Y=Dropout(ReLU(Y_{attn}
 ))
\end {equation}
\begin{figure*}
    \centering
    \includegraphics[width=0.8\linewidth]{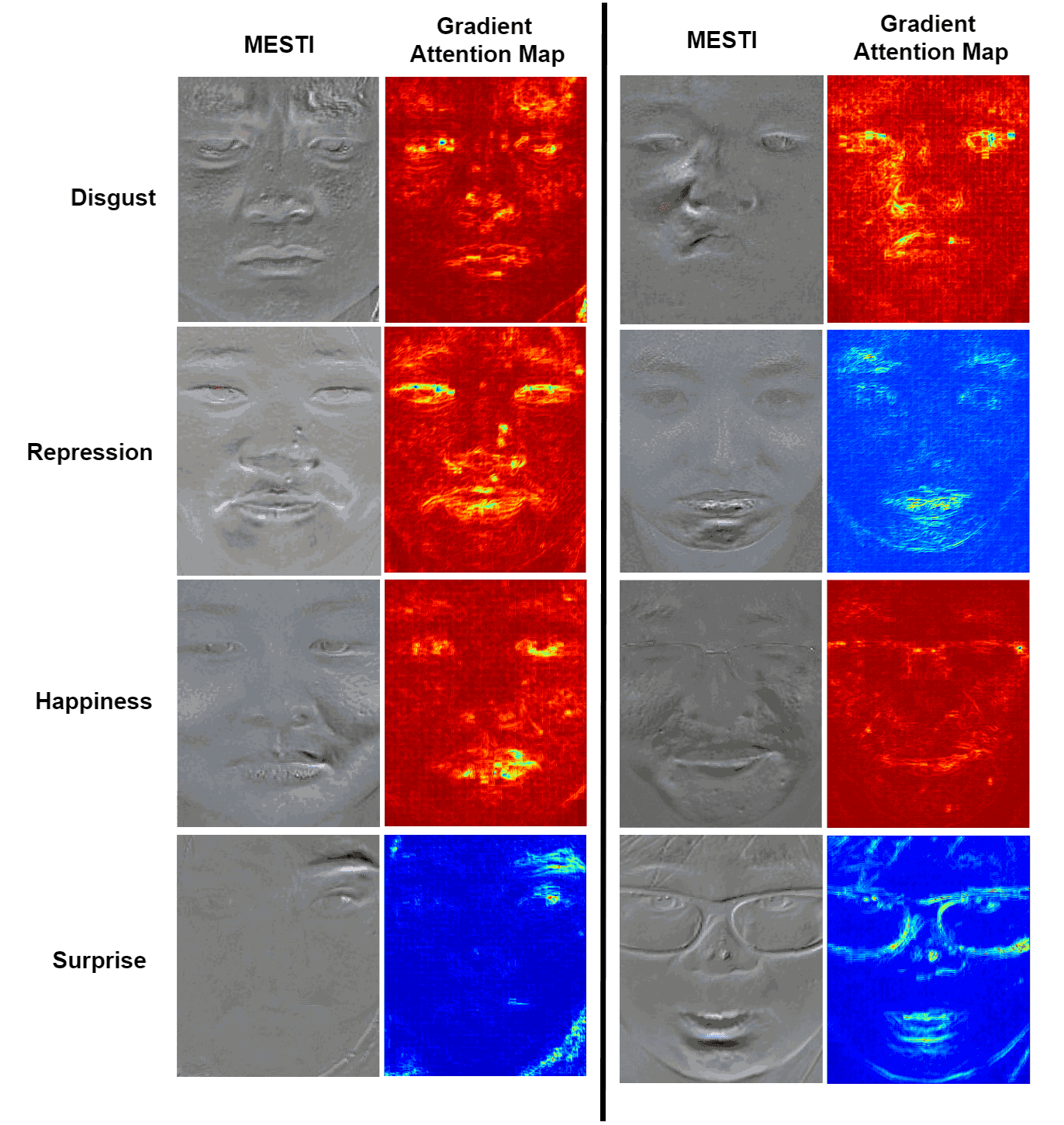}
    \caption{Visualization of MESTI and corresponding Gradient Attention Map characterize each emotion of ME (Best viewed in color)}
    \label{fig:enter-label}
\end{figure*}

\section{EXPERIMENTS AND RESULTS}
\subsection{Experiment scenarios and objectives}
To evaluate the effectiveness of the proposed method for MER, which includes the MESTI as input representation, the MEGANet as MER network, and the combined approach of MESTI and MEGANet, three experimental scenarios were conducted to assess the performance of each proposed component:

\textbf{Experiment 01:} This experiment aims to evaluate the effectiveness of the MESTI input representation. Specifically, it compares MESTI with other input modalities previously used in MER studies, such as Apex Frame, Optical Flow, Dynamic Image, Active Image, and Affective Image. Furthermore, the experiment continues by replacing the input in previously published MER networks with MESTI to investigate whether MESTI improves MER performance in these prior works.

\textbf{Experiment 02:} This experiment evaluates the performance of MEGANet in MER. And analysis the effective of key block proposed in MEGANet.

\textbf{Experiment 03:} This experiment assesses the overall effectiveness of the proposed MER method, combining MESTI and MEGANet. The results of this experiment are compared with recent SOTA methods to demonstrate the superiority of the proposed approach.

\textbf{Experiment 04:} This experiment assesses the generalization of the proposed MER method. The results of this experiment are compared with recent SOTA methods in cross-dataset protocol to demonstrate the superiority of the proposed approach.

\subsection{Dataset \& Data preprocessing}
\subsubsection{Datasets}
The experiments are conducted on three publicly available ME recognition datasets, namely SMIC-HS\cite{6553717}, CASME II\cite{Yan2014-fa} and SAMM\cite{7492264}, which are widely used as standard benchmarks for MER and for comparison with previous studies.

\subsubsection{Data preprocessing} To ensure a fair comparison, the data preprocessing steps are minimalized, limiting them to face cropping, histogram equalization and resizing the images to dimensions appropriate for each network’s input requirements. This minimalistic approach eliminates potential biases from complex preprocessing techniques, allowing us to isolate and highlight the contributions of each input modality to overall network performance.

To ensure fair comparison with prior studies, this work conducts both 3-class and 5-class evaluations.
In the 3-class evaluation, three common categories across the datasets are considered: positive, negative, and surprise. For the 5-class evaluation, the original emotion annotations provided in the CASME II and SAMM datasets are retained. Specifically, the CASME II dataset comprises the ME labels disgust (60 samples), happiness (33), other (102), repression (27), and surprise (25). The SAMM dataset includes the labels anger (57 samples), happiness (26), contempt (12), surprise (15), and other (49).

\subsection{Experimental settings}
\subsubsection{Experiment 01} To ensure a fair comparison of the effectiveness of all input modalities, a common procedure is applied to all modalities. A train-test split protocol is used, with 90\% for training and 10\% for testing. The input is sequentially fed into three widely recognized deep learning networks: VGG19, ResNet50, and EfficientNetB0. This standardized approach minimizes external factors that could influence performance outcomes, allowing the observed differences to be directly attributed to the input modality itself.

MESTI is further used as an alternative input for the MER networks employed in two prior studies. To ensure fairness and the significance of the comparison results, we implement the experimental method in the same manner as described in their studies. Both studies used the Leave-One-Subject-Out (LOSO) protocol for evaluation.
\begin{table*}[t]
\centering
\caption{Comparison of input modalities in MER with Train–Test split protocol.}
\label{tab:input_modalities}
\setlength{\tabcolsep}{1pt}
\begin{tabular}{llcc|cc|cc}
\toprule
\multicolumn{2}{c}{\textbf{Input modality for MER}} &
\multicolumn{2}{c|}{\textbf{VGG19}} &
\multicolumn{2}{c|}{\textbf{ResNet50}} &
\multicolumn{2}{c}{\textbf{EfficientNet-B0}} \\
\cmidrule(lr){3-4}\cmidrule(lr){5-6}\cmidrule(l){7-8}
 & & \textbf{CASME II} & \textbf{SAMM} & \textbf{CASME II} & \textbf{SAMM} & \textbf{CASME II} & \textbf{SAMM} \\
\midrule
\multicolumn{8}{l}{\emph{Static}} \\
\midrule
& Apex frame
& 50.00\% & 43.75\% & 46.15\% & 50.00\% & 34.62\% & 43.75\% \\
\midrule
\multicolumn{8}{l}{\emph{Dynamic}} \\
\midrule
& Optical flow (Onset–Apex)
& 50.00\% & 50.00\% & 46.15\% & 37.50\% & 26.92\% & 43.75\% \\
& Optical flow (Apex–Offset)
& 53.85\% & 50.00\% & 34.62\% & 43.75\% & 46.15\% & 43.75\% \\
& Dynamic image
& 57.69\% & 50.00\% & 53.85\% & 37.50\% & 53.85\% & \textbf{50.00\%} \\
& Affective motion image
& 50.00\% & 43.75\% & 53.85\% & 50.00\% & 46.15\% & 43.75\% \\
& Active image
& 48.00\% & 57.14\% & 44.00\% & 50.00\% & 52.00\% & 48.00\% \\
& \textbf{MESTI (ours)}
& \textbf{73.08\%} & \textbf{62.50\%} & \textbf{65.38\%} & \textbf{56.25\%} & \textbf{61.54\%} & \textbf{50.00\%} \\
\bottomrule
\end{tabular}
\end{table*}

\begin{table}[t]
\centering
\caption{MESTI Input representation in published MER networks with LOSO protocol compared with the original research.}
\label{tab:acc_inputs_networks}
\begin{threeparttable}
\setlength{\tabcolsep}{8pt}
\begin{tabular}{llcc}
\toprule
\multirow{2}{*}{Input} & \multirow{2}{*}{Network} & \multicolumn{1}{c}{CASME II} & \multicolumn{1}{c}{SAMM} \\
 &  & ACC & ACC \\
\midrule
Apex Image\textsuperscript{*}   & Micro-attention \cite{wang2020micro} & 65.9\%  & 48.5\%  \\
MESTI                            & Micro-attention \cite{wang2020micro} & \textbf{71.02\%} & \textbf{63.24\%} \\
\midrule
Dynamic Image\textsuperscript{*} & VGG19 \cite{8844867} & 51.02\% & 43.23\% \\
MESTI                             & VGG19 \cite{8844867} & \textbf{69.39\%} & \textbf{60.29\%} \\
\bottomrule
\end{tabular}

\begin{tablenotes}[flushleft]
\footnotesize
\item[\textsuperscript{*}] Results from the original research (baseline input); other rows use our MESTI representation.
\end{tablenotes}
\end{threeparttable}
\end{table}
\subsubsection{Experiment 02} MEGANet is evaluated through an ablation study. Two ablation scenarios are conducted. In the first, individual components of MEGANet: the Gradient Attention Block and the Residual Attention Block are isolated and evaluated independently. In the second, the performance of MEGANet is assessed with respect to the varying number of Residual Attention Blocks to determine the most suitable configuration. Both ablations used the LOSO protocol for evaluation.

\subsubsection{Experiment 03} This experiment is designed to evaluate the proposed method in this paper, using MESTI as the input and MEGANet as the MER network. The experiment is conducted using the LOSO protocol for evaluation to ensure a meaningful comparison with previously published methods.

\subsubsection{Experiment 04} This experiment is designed to evaluate the proposed method in cross-dataset protocol, using MESTI as the input and MEGANet as the MER network. The experiment is conducted using the CASMEII/SAMM for trainning and testing on the SMIC dataset, evaluate on ACC and WF1 metric.

\subsubsection{Specific configuration and training methodology} The following configuration and training methodology were used in this study:
\begin{itemize}
    \item Data Augmentation: The dataset is augmented using horizontal flipping and rotations at 5° and 10° (both clockwise and counterclockwise).
    \item Loss Function: Focal Loss was used to address class imbalance and improve the network's focus on hard-to-classify samples.
    \item Optimizer: The Adam optimizer was employed with a learning rate of $1e-4$ and weight decay of $1e-5$ to optimize the network, no learning-rate scheduler was used.
    \item Training Duration: The network was trained for 50 epochs.
    \item Metric: The primary evaluation metrics are UF1 and UAR; accuracy is reported as a supplementary measure, WF1 used in cross dataset evaluation follow other previous methods \cite{shao2025mol}.
\end{itemize}

\subsection{Results}
\subsubsection{\textbf{Visual representation}}
The visual results of MESTI and its corresponding Gradient Attention Map are shown in Figure 8 to observe how MESTI captures the characteristic features of each ME emotion type and how the Gradient Attention Map highlights the regions of interest within MESTI. A key observation is that MESTI effectively captures and highlights the defining motion patterns of MEs, making them perceptible to the human eye in a single image representation.

More specifically, both MESTI and the Gradient Attention Map successfully depict the characteristic Action Units corresponding to different ME emotions. For Disgust, the key motion regions primarily appear around the eyebrows, one side of the nose, and the corners of the mouth. Repression manifests as subtle downward movements on both sides of the mouth and the chin. Happiness is expressed by an upward motion at the corners of the mouth, whereas Surprise is predominantly reflected in eyebrow elevation and lower lip movement. These findings highlight the capability of MESTI to encode motion dynamics effectively in a compact and visually interpretable format.
\begin{table*}[t]
\centering
\caption{Comparison with recent SOTA methods in 3-class evaluation with LOSO protocol}
\label{tab:me_results}
\begin{threeparttable}
\setlength{\tabcolsep}{2pt}
\begin{tabular}{llccc|ccc|ccc}
\toprule
\multirow{2}{*}{Method} & \multirow{2}{*}{Year} &
\multicolumn{3}{c|}{CASME II} &
\multicolumn{3}{c|}{SMIC-HS} &
\multicolumn{3}{c}{SAMM} \\
\cmidrule(lr){3-5}\cmidrule(lr){6-8}\cmidrule(l){9-11}
 & & UF1 & UAR & ACC & UF1 & UAR & ACC & UF1 & UAR & ACC \\
\midrule
FeatRef \cite{ZHOU2022108275} & 2022 & 0.892 & 0.887 & -- & 0.701 & 0.708 & -- & 0.737 & 0.716 & -- \\
Dual-ATME \cite{e25030460} & 2023 & 0.765 & 0.751 & 0.817 & 0.646 & 0.658 & 0.646 & 0.562 & 0.538 & 0.714 \\
FRL-DGT \cite{zhai2023featurerepresentationlearningadaptive} & 2023 & 0.919 & 0.903 & -- & 0.743 & 0.749 & -- & \underline{0.772} & \underline{0.758} & -- \\
SelfME \cite{10204934} & 2023 & 0.908 & \underline {0.929} & -- & 0.697 & 0.701 & -- & -- & -- & -- \\
Micron-BERT \cite{Nguyen2023MicronBERTBF} & 2023 & 0.903 & 0.891 & -- & 0.855 & 0.838 & -- & -- & -- & -- \\
MERASTC \cite{9363624} & 2023 & \textbf{0.933} & \textbf{0.950} & -- & 0.790 & 0.862 & -- & -- & -- & -- \\
GLEFFN \cite{10.1145/3607829.3616446} & 2023 & 0.883 & 0.911 & -- & 0.771 & 0.786 & -- & -- & -- & -- \\
SODA4MER \cite{soda4mer} & 2025 & 0.887 & 0.881 & -- & \underline{0.886} & \underline{0.888} & -- & -- & -- & -- \\
OFVIG-Net \cite{ofvig} & 2025 & 0.713 & 0.720 & -- & 0.644 & 0.640 & -- & 0.607 & 0.579 & -- \\
\textbf{MESTI-MEGANet} & 2025 & 0.913 & \underline{0.929} & 0.932 & \textbf{0.917} & \textbf{0.924} & \textbf{92.68} & \textbf{0.890} & \textbf{0.914} & \textbf{0.918} \\
\bottomrule
\end{tabular}
\begin{tablenotes}[flushleft]
\footnotesize
\item \textbf{Bold} indicates the best result in each column; \underline{underline} indicates the second-best result.
\item “–” denotes that the metric was not reported in the cited work.
\end{tablenotes}
\end{threeparttable}
\end{table*}

\begin{table*}[t]
\centering
\caption{Comparison with recent SOTA methods in 5-class evaluation with LOSO protocol.}
\label{tab:results_3metrics}
\begin{threeparttable}
\setlength{\tabcolsep}{6pt}
\begin{tabular}{llccc|ccc}
\toprule
\multirow{2}{*}{Method} & \multirow{2}{*}{Year} &
\multicolumn{3}{c|}{CASME II} &
\multicolumn{3}{c}{SAMM} \\
\cmidrule(lr){3-5}\cmidrule(l){6-8}
 & & UF1 & UAR & ACC & UF1 & UAR & ACC \\
\midrule
GEME \cite{NIE202113} & 2021 & 0.735 & - & 75.20 & 0.454 & - & 55.88 \\
MER-Supcon \cite{mersupcon} & 2022 & 0.729 & - & 73.58 & 0.625 & - & 67.65 \\
CMNet \cite{wei2023cmnet} & 2023 & 0.740 & - & 78.05 & 0.772 & - & 78.68 \\
C3DBed \cite{PAN2023106258} & 2023 & 0.752 & - & 77.64 & 0.722 & - & 75.73 \\
KPCANet \cite{kpcanet} & 2023 & 0.659 & - & 70.46 &  0.522 & - & 63.83 \\
JGULF \cite{WANG2024105091}   & 2024  & \underline{0.807} & - & \underline{82.04} & 0.720 & - & \underline{80.71} \\
AU GCN \cite{10446702}   & 2024  & 0.776 & - & 81.85 & 0.757 & - & 79.82 \\
SODA4MER \cite{soda4mer}   & 2025  & \textbf{0.814} & - & \textbf{84.18} &  \underline{0.789} & - & 80.30 \\
LRT3O \cite{10981613} & 2025 & 0.791 & - & 81.78 & 0.757 & - & 80.15 \\
Micro\_NesT \cite{HE2025128372} & 2025 & 0.772 & - & 77.93 & 0.748 & - & 76.69 \\
MELLM \cite{mellm} & 2025 & 0.485 & \underline{0.534} & 64.34 & - & - & - \\
\textbf{MESTI-MEGANet} & 2025 & 0.779 & \textbf{0.786} & \underline{82.04} & \textbf{0.791} & \textbf{0.803} & \textbf{80.88} \\
\bottomrule
\end{tabular}

\begin{tablenotes}[flushleft]
\footnotesize
\item \textbf{Bold} indicates the best result in each column; \underline{underline} indicates the second-best result.
\item “–” denotes that the metric was not reported in the cited work.
\end{tablenotes}
\end{threeparttable}
\end{table*}

\begin{table}[t]
\centering
\caption{Comparison with state-of-the-art methods in terms of cross-dataset evaluation with 3-classes. The ACC and WF1 metric used for compare with SOTA methods}
\label{tab:acc_inputs_networks}
\begin{threeparttable}
\setlength{\tabcolsep}{8pt}
\begin{tabular}{lc|cc|cc}
\toprule
\multirow{2}{*}{Method} & \multirow{2}{*}{Year} & \multicolumn{2}{c|}{CASME II $\rightarrow$ SMIC} & \multicolumn{2}{c}{SAMM $\rightarrow$ SMIC} \\
 & & Acc & WF1 & Acc & WF1 \\
\midrule
STCNN \cite{8852419} & 2019 & 31.40 & 19.00 & 32.50 & 19.00 \\
CapsuleNet \cite{10.1109/FG.2019.8756544}& 2019 & 32.20 & 15.20 & 32.40 & 17.90 \\
MER-GCN \cite{9175230}  & 2020 & 36.70 & 27.20 & 36.10 & 17.80 \\
AU-GACN \cite{auasis} & 2020 & 34.40 & 31.90 & \underline{45.10} & 30.90 \\
MOL \cite{shao2025mol} & 2025 & \underline{47.13} & \underline{43.91} & 44.58 & \underline{32.32} \\
\textbf{Ours} & 2025 & \textbf{50.00} & \textbf{46.81} & \textbf{46.95} & \textbf{40.89} \\
\bottomrule
\end{tabular}
\begin{tablenotes}[flushleft]
\footnotesize
\item \textbf{Bold} indicates the best result in each column; \underline{underline} indicates the second-best result.
\item CASMEII $\rightarrow$ SMIC denotes training on CASMEII and testing on SMIC.
\end{tablenotes}
\end{threeparttable}
\end{table}

\subsubsection{\textbf{MESTI Representation compared with other input modalities}}
Table \ref{tab:input_modalities} summarizes the comparative performance of various input modalities in the ME recognition task, evaluated using deep learning network on the CASMEII and SAMM datasets. The results consistently demonstrate that MESTI outperforms all other input modalities across the three widely used CNN architectures: VGG19, ResNet50, and EfficientNetB0. Specifically, MESTI achieves the highest accuracy of 73.08\% on CASMEII and 62.5\% on SAMM with VGG19, surpassing the second-best input modality (Dynamic Image) by 15.39\% and 12.5\%, respectively. This superior performance underscores MESTI’s effectiveness in capturing subtle motion features, which are crucial for ME recognition.

For the SAMM dataset, the overall recognition performance is lower compared to CASMEII across all input modalities, a trend consistent with previous studies due to SAMM’s greater diversity and complexity. Despite this challenge, MESTI continues to demonstrate superior recognition capabilities, achieving 62.5\% with VGG19 and 56.25\% with ResNet50, reinforcing its robustness across different datasets and deep learning architectures.

To further validate MESTI’s effectiveness, we investigated whether its superior performance was specific to our proposed pipeline or if it could enhance other established MER architectures. The original input modalities are replaced by two previously published works with MESTI: VGG19 (originally using Dynamic Image) and Micro-Attention (originally using Apex Frame). The results, presented in Table 2, show that for VGG19, replacing the input with MESTI improved recognition accuracy from 51.02\% to 69.39\% on CASMEII and from 43.23\% to 60.29\% on SAMM. Similarly, for Micro-Attention, using MESTI as input improved accuracy from 65.90\% to 71.02\% on CASMEII and from 48.5\% to 63.24\% on SAMM. These results confirm that MESTI not only enhances our proposed network but also significantly improves the performance of other MER architectures, demonstrating its capability to effectively represent ME dynamics in a single image.

\begin{table*}[t]
\centering
\caption{Ablation study of key blocks of MEGANet.}
\label{tab:soda4mer_ablation}
\setlength{\tabcolsep}{10pt}
\begin{tabular}{c|c|c|ccc}
\toprule
Gradient Attention Block & Residual Block & Self-attention & UF1 & UAR & ACC \\
\midrule
-- & \checkmark & -- & 0.788 & 0.821 & 80.49 \\
-- & -- & \checkmark & 0.746 & 0.775 & 75.61 \\
-- & \checkmark & \checkmark & 0.8304 & 0.8609 & 83.54 \\
\checkmark & -- & -- & 0.8084 & 0.8551 & 81.10 \\
\checkmark & \checkmark & \checkmark & \textbf{0.917} & \textbf{0.924} & \textbf{92.68} \\
\bottomrule
\end{tabular}
\end{table*}


\begin{table}[t]
\centering
\caption{Ablation study of number of Residual Attention Blocks in SMIC dataset}
\label{tab:ma_gamma}
\setlength{\tabcolsep}{10pt}
\begin{tabular}{c|ccc}
\toprule
Residual Attention Block & UF1 & UAR & ACC \\
\midrule
$\times$ 2 & 0.802 & 0.842 & 82.32 \\
$\times$ 3 & \textbf{0.917} & \textbf{0.924}  & \textbf{92.68} \\
$\times$ 4 & 0.844 & 0.878 & 85.98 \\
\bottomrule
\end{tabular}
\end{table}
\subsubsection{\textbf{Compared with State-of-the-art methods in MER}}
The comparative results with recent state-of-the-art methods are reported in Table 3 (for the 3-class evaluation), Table 4 (for the 5-class evaluation) and Table 5 (for cross-dataset evaluation). Overall, the proposed method outperforms existing SOTA approaches on the SAMM and SMIC-HS datasets and achieves competitive performance on the CASME II dataset, as reflected across all three evaluation metrics: accuracy, UF1, and UAR.\\\\
\textbf{3-class evaluation}\\
Overall, the proposed MESTI-MEGANet demonstrates strong and balanced performance across the three benchmark datasets, achieving the best results on SMIC-HS and SAMM, while remaining competitive on CASME II.

On SMIC-HS, the proposed method achieves the best performance across all three reported metrics, with UF1 = 0.917, UAR = 0.924, and ACC = 92.68\%. Compared with the strongest competing method in this table, SODA4MER, which reports UF1 = 0.886 and UAR = 0.888, our method improves by +0.031 in UF1 and +0.036 in UAR. These results indicate that the proposed framework is particularly effective on SMIC-HS, even though this dataset does not provide apex-frame annotations.

On SAMM, MESTI-MEGANet again ranks first on all three metrics, reaching UF1 = 0.890, UAR = 0.914, and ACC = 0.918. Since several competing methods do not report results on SAMM, the comparison is limited to the available entries in Table 3; nevertheless, among the reported methods, our approach shows the strongest overall performance. In particular, it surpasses FRL-DGT (UF1 = 0.772, UAR = 0.758) and Dual-ATME (UF1 = 0.562, UAR = 0.538, ACC = 0.714) by a clear margin, demonstrating the effectiveness of the proposed representation and architecture on this challenging dataset.

On CASME II, the proposed method remains competitive, obtaining UF1 = 0.913, UAR = 0.929, and ACC = 0.932. The best performance on this dataset is achieved by MERASTC, with UF1 = 0.933 and UAR = 0.950, while FRL-DGT reports UF1 = 0.919 and UAR = 0.903. Therefore, although our method does not rank first on CASME II, the performance gap to the strongest reported methods remains small, indicating that the proposed framework generalizes well across different benchmark conditions.

Taken together, the results in Table 3 show that MESTI-MEGANet achieves state-of-the-art performance on SMIC-HS and SAMM in the 3-class setting, while maintaining highly competitive performance on CASME II. This suggests that the combination of MESTI, which provides an apex-centered motion representation, and MEGANet, which explicitly enhances subtle motion-sensitive regions, is effective across datasets with different characteristics and recording conditions.\\\\
\textbf{5-class evaluation}\\
On CASMEII, SODA4MER yields the best UF1 and ACC (UF1=0.814, ACC=84.18\%). MESTI-MEGANet reaches UF1=0.779, UAR=0.786, ACC=82.04 and matches JGULF on accuracy (82.04), trailing SODA4MER by 2.14.
On SAMM, MESTI-MEGANet attains the top results across all reported metrics (UF1=0.791, UAR=0.803, ACC=80.88). The gains are small but consistent: UF1 is slightly higher than SODA4MER (0.789), and accuracy exceeds JGULF (80.71) by 0.17 and SODA4MER (80.30) by 0.58. Note that UAR on SAMM is not commonly reported by most baselines, so direct UAR comparisons are limited.

Across protocols and datasets, MESTI-MEGANet delivers  SOTA on SMIC-HS (3-class) and strong SOTA on SAMM (both 3-class and 5-class), while remaining competitive on CASMEII (second-best accuracy in 3-class; tied for accuracy but below the best UF1 in 5-class). These outcomes indicate that the method generalizes well to different datasets and label granularities, with the largest margins observed on SMIC-HS (dataset without apex frame annotated).

This success is attributed to two key factors:
\begin{itemize}
    \item MESTI’s motion-specific encoding, which preserves spatiotemporal dynamics (Figure 2)
    \item MEGANet’s Gradient Attention mechanism, which focuses on intensity transitions (Figure 5) while Residual Attention blocks model long-range dependencies (Figure 6).\\
\end{itemize}
\textbf{Cross-dataset evaluation}\\
To further assess the robustness of the proposed method under domain shift, we additionally evaluate MESTI-MEGANet in a cross-dataset setting and compare it with representative state-of-the-art methods, as shown in Table 5. In this protocol, the model is trained on one dataset and tested on another, which is substantially more challenging than within-dataset LOSO evaluation due to differences in subjects, recording conditions, elicitation procedures, and data distributions. Following prior cross-dataset MER studies, ACC and WF1 are reported here to enable fair comparison with published baselines.

The results show that the proposed method achieves the best performance in all reported transfer settings. Specifically, when trained on CASME II and tested on SMIC, our method reaches 50.00\% ACC and 46.81\% WF1, outperforming the previous best method MOL (47.13\% ACC, 43.91\% WF1) by +2.87 and +2.90 percentage points, respectively. Similarly, in the more challenging SAMM → SMIC transfer, MESTI-MEGANet achieves 46.95\% ACC and 40.89\% WF1, again surpassing MOL (44.58\% ACC, 32.32\% WF1) by +2.37 points in accuracy and a larger margin of +8.57 points in WF1. These improvements are particularly meaningful because WF1 is more informative in imbalanced settings and indicates that the gain is not limited to majority-class prediction.

Overall, these cross-dataset results suggest that the proposed framework is not only effective under standard within-dataset evaluation, but also exhibits improved robustness when the training and testing distributions differ. We attribute this behavior to two complementary factors. First, MESTI provides a compact motion-centered representation that preserves discriminative micro-expression dynamics while reducing sensitivity to raw-frame appearance variation. Second, MEGANet, especially its Gradient Attention Block, helps focus the model on motion-sensitive facial patterns that are more likely to transfer across datasets than dataset-specific appearance cues. Although the absolute performance in cross-dataset evaluation remains lower than LOSO results, which is expected in MER, the consistent gains over prior methods indicate stronger generalization ability under domain shift.

\begin{table}[t]
\centering
\caption{Ablation study of the attention mechanism in MEGANet in cross-dataset evaluation}
\label{tab:acc_inputs_networks}
\begin{threeparttable}
\setlength{\tabcolsep}{8pt}
\begin{tabular}{l|cc|cc}
\toprule
\multirow{2}{*}{Attention type} & \multicolumn{2}{c|}{CASME II $\rightarrow$ SMIC} & \multicolumn{2}{c}{SAMM $\rightarrow$ SMIC} \\
 & Acc & WF1 & Acc & WF1 \\
\midrule
None  & 45.12 & 44.67 & 39.63 & 39.97 \\
SE \cite{hu2018squeeze} & 44.51 & 42.10 & 43.90 & 40.74 \\
CBAM \cite{woo2018cbam}   & 45.73 & 43.15 & 41.46 & 39.28 \\
\textbf{Gradient Attention}  & \textbf{50.00} & \textbf{46.81} & \textbf{46.95} & \textbf{40.89} \\
\bottomrule
\end{tabular}
\end{threeparttable}
\end{table}

\subsubsection{\textbf{The dependency of MESTI on apex frame}}
\begin{table}[t]
\centering
\caption{Ablation study of the dependency of apex frame of MESTI, comparision with other input representations in cross-dataset evaluation}
\label{tab:acc_inputs_networks}
\begin{threeparttable}
\setlength{\tabcolsep}{8pt}
\begin{tabular}{l|cc|cc}
\toprule
\multirow{2}{*}{Input representation} & \multicolumn{2}{c|}{CASME II $\rightarrow$ SMIC} & \multicolumn{2}{c}{SAMM $\rightarrow$ SMIC} \\
 & Acc & WF1 & Acc & WF1 \\
\midrule
Apex frame  & 35.98 & 30.04 & 39.63 & 37.36 \\
Dynamic image & 40.85 & 40.08 & 40.24 & 36.09 \\
MESTI-middle based  & 48.17 & 42.50 & 46.34 & 40.83 \\
\textbf{MESTI-apex based}  & \textbf{50.00} & \textbf{46.81} & \textbf{46.95} & \textbf{40.89} \\
\bottomrule
\end{tabular}
\end{threeparttable}
\end{table}

In the SMIC dataset, apex frame annotations are not provided; hence, the apex frame information cannot be directly utilized to construct MESTI. As an alternative, in this study we adopt a simple strategy of selecting the middle frame.

Interestingly, the results on SMIC demonstrate strong performance despite the absence of apex frame supervision. This finding suggests that the proposed MESTI approach can remain effective even without precise apex frame information, highlighting its robustness and applicability in more challenging scenarios where apex annotations are unavailable.

We continue to expand our experiments to analyze performance under noisy apex or apex-absent conditions, evaluated within a cross-dataset setting. This experiment serves two objectives: first, to compare the impact of using precisely labeled apex frames versus substituting them with middle frames; and second, to evaluate the effectiveness of different inputs to investigate MESTI's performance in cross-dataset scenarios. The results are recorded in Table 9.

Table 9 provides two complementary observations. First, the performance of MESTI is indeed affected by the choice of apex location, confirming that apex-centered temporal encoding is relevant to the quality of the representation. However, the degradation remains relatively limited when the exact apex is replaced by an approximate one, indicating that MESTI is not overly sensitive to small or moderate inaccuracies in apex localization. In other words, while precise apex information is beneficial, the proposed representation does not collapse when the apex is noisy or unavailable.

Second, even under the apex-free setting, MESTI remains competitive with, and in several cases superior to, the other input modalities. This suggests that the advantage of MESTI does not rely solely on having perfectly annotated apex frames, but also comes from its ability to convert the facial motion sequence into an apex-centered spatio-temporal representation that still preserves discriminative motion structure under approximate temporal centering.

These results further support the practical robustness of the proposed method. In real-world MER scenarios, accurate apex annotations may be unavailable or difficult to estimate reliably. The findings in Table 9 suggest that MESTI can still provide an effective representation under such conditions, with only a limited performance drop relative to the exact-apex case. Therefore, the proposed formulation is not restricted to ideal benchmark settings, but remains applicable in more realistic apex-free or noisy-apex situations.

\subsection{Ablation study}
\textbf{Ablation on network architecture.} To evaluate the contribution of each component in MEGANet, we conducted an ablation study focusing on both the key building blocks and the number of Residual Attention Blocks.

As shown in Table 6, removing any of the major components leads to a clear performance drop. Using only the Residual Block without Gradient Attention or Self-attention yields the lowest performance (UF1 = 0.746, UAR = 0.775, ACC = 75.61). Incorporating Self-attention alone provides some improvement (ACC = 83.54), while Gradient Attention Block alone achieves ACC = 81.10. The best performance is obtained when all three modules are integrated, resulting in significant gains (UF1 = 0.917, UAR = 0.924, ACC = 92.68). This demonstrates that the Gradient Attention Block, Residual Block, and Self-attention contribute complementary benefits, and their combination is essential for maximizing recognition accuracy.


Table 7 further analyzes the effect of network depth by varying the number of Residual Attention Blocks. The results show that the performance improves markedly when the depth increases from two to three blocks (in term of ACC, UF1, UAR increase from 82.32/0.802/0.842 to 92.68/0.917/0.924), but decreases when a fourth block is added (85.98/0.844/0.878 in ACC, UF1, UAR). We interpret this behavior as a trade-off between representational capacity and generalization. With only two blocks, the network may not be sufficiently deep to progressively refine the subtle motion cues emphasized by the Gradient Attention Block and to model higher-level spatial dependencies. Three blocks provide a stronger hierarchical refinement process and yield the best overall performance. However, increasing the depth further to four blocks introduces additional model complexity, which is less suitable for MER due to the limited size and high subject variability of current datasets. In this setting, the additional block is more likely to introduce redundancy or overfitting than further discriminative benefit. Therefore, three Residual Attention Blocks provide the most effective balance between feature refinement and robustness in our experiments.\\\\
\textbf{Ablation on attention mechanism.} To further validate the effectiveness of the proposed Gradient Attention Block, we compare it with several representative attention alternatives, including no attention, SE, and CBAM, under the same cross-dataset evaluation setting. The results are summarized in Table 8.

The proposed Gradient Attention consistently achieves the best performance across both transfer scenarios. On CASME II $\rightarrow$ SMIC, it obtains 50.00\% ACC and 46.81\% WF1, outperforming the no-attention baseline (45.12\% ACC, 44.67\% WF1), SE (44.51\% ACC, 42.10\% WF1), and CBAM (45.73\% ACC, 43.15\% WF1). On SAMM → SMIC, Gradient Attention again ranks first with 46.95\% ACC and 40.89\% WF1, compared with 39.63/39.97 for no attention, 43.90/40.74 for SE, and 41.46/39.28 for CBAM. These results indicate that the gain is not simply due to adding an arbitrary attention module, but is specifically related to the proposed gradient-driven formulation.

A explanation is that generic attention mechanisms such as SE and CBAM mainly reweight channels or spatial regions from feature activations, whereas the proposed Gradient Attention Block derives the attention cue directly from local intensity transitions. This is particularly suitable for MER, because the discriminative signal in MESTI is expressed as subtle motion-induced contrast changes rather than strong semantic structures. Therefore, Gradient Attention provides a more effective inductive bias for highlighting motion-relevant regions, especially in the cross-dataset setting where robustness to appearance variation is critical.

Another notable observation is that the advantage of Gradient Attention becomes more pronounced in the SAMM $\rightarrow$ SMIC setting, where the domain gap is larger. This suggests that emphasizing gradient-based motion traces may help the model rely less on dataset-specific appearance patterns and more on transferable motion-sensitive cues. These findings support the claim that the proposed block is not merely a generic attention add-on, but a representation-aware mechanism that is well matched to the motion structure encoded by MESTI.

\subsection{DISCUSSION AND LIMITATIONS}
\noindent\textit{\textbf{Class imbalance.}} MER datasets are inherently imbalanced, especially in the 5-class setting, as reflected by the label distributions of CASME II and SAMM reported in Section IV-B. In this work, class imbalance was handled at multiple levels. At the optimization level, we used Focal Loss to reduce the dominance of easy majority-class samples and encourage the model to focus more on difficult and minority-class examples. At the evaluation level, we emphasized UF1 and UAR in the main comparative and ablation experiments, since these metrics are more informative than accuracy for imbalanced MER datasets. In addition, moderate data augmentation was used to improve robustness and reduce overfitting to dominant classes.\\\\
\textit{\textbf{Dependency on apex-frame estimation.}} MESTI uses apex information to define the center of its apex-oriented ranking formulation. However, the method is not strictly dependent on perfectly accurate apex detection. Its goal is to concentrate representational emphasis around the most informative phase of the micro-expression rather than to require exact frame-level precision. This property is partially supported by the SMIC-HS setting, where apex annotations are unavailable and a simple middle-frame approximation is used, yet the proposed framework still achieves strong performance. This suggests that MESTI can remain effective when the apex is unavailable or only approximately estimated. Nevertheless, large errors in apex localization may shift the temporal weighting away from the truly discriminative phase and thus weaken the resulting representation.\\\\
\textit{\textbf{Failure cases and limitations.}} Despite its effectiveness, the proposed framework still has several limitations. First, MESTI compresses an entire video into a single image, which improves compactness but may lose some fine-grained temporal ordering information. Second, the quality of the representation may degrade when the estimated apex is substantially inaccurate. Third, strong head motion, motion blur, occlusion, or illumination variation may still interfere with the encoding of subtle facial motion, especially when the ME itself is extremely weak. Fourth, due to the small scale and high subject variability of current MER datasets, deeper or more complex architectures can easily overfit, which is also consistent with our observation that increasing the number of Residual Attention Blocks beyond three does not improve performance.

\section{CONCLUSION}
In this work, we address the limitations of existing MER methodologies by introducing ME Spatio-Temporal Image as a novel input modality and ME Gradient Attention Network as a novel architecture. MESTI effectively encodes micro-movements into a single image, preserving both spatial and temporal features, while MEGANet utilizes a Gradient Attention mechanism to enhance the detection of subtle motion cues.

Our experimental results validate the effectiveness of MESTI by showing that it outperforms all other input modalities, including Apex Frame, Optical Flow, and Dynamic Image, across multiple deep learning networks. Furthermore, replacing the input of previously published MER architectures with MESTI results in significant improvements in recognition accuracy, highlighting its broad applicability. Additionally, MEGANet achieves state-of-the-art performance, particularly when combined with MESTI, confirming its effectiveness in ME analysis. These findings establish MESTI and MEGANet as highly effective solutions for MER, significantly improving recognition accuracy. Future work could explore refining MESTI for real-time applications, integrating additional attention mechanisms, or leveraging larger-scale datasets to further advance ME recognition systems.

\bibliographystyle{cas-model2-names}

\bibliography{cas-refs}

@article{Ekman01051992,
  author = {Ekman, P.},
  year = {1992},
  title = {An argument for basic emotions},
  journal = {Cognition And Emotion},
  volume = {6},
  pages = {169-200},
}

@article{Yan2013,
  author = {Yan, W. and Wu, Q. and Liang, J. and Chen, Y. and Fu, X.},
  title = {How Fast are the Leaked Facial Expressions: The Duration of Micro-Expressions},
  journal = {Journal Of Nonverbal Behavior},
  volume = {37},
  pages = {217-230},
  note = {(2013,12)},
  doi = {10.1007/s10919-013-0159-8},
}

@inproceedings{8852419,
  author = {Teja Reddy, S. and Teja Karri, S. and Dubey, S. and Mukherjee, S.},
  year = {2019},
  title = {Spontaneous Facial Micro-Expression Recognition using 3D Spatiotemporal Convolutional Neural Networks},
  booktitle = {2019 International Joint Conference On Neural Networks (IJCNN)},
  pages = {1-8},
}

@article{Matsumoto2011,
  author = {Matsumoto, D. and Hwang, H.},
  title = {Evidence for training the ability to read microexpressions of emotion},
  journal = {Motivation And Emotion},
  volume = {35},
  pages = {181-191},
  note = {(2011,6)},
  doi = {10.1007/s11031-011-9212-2},
}

@article{Ekman2003-eb,
  author = {Ekman, P.},
  title = {Darwin, deception, and facial expression},
  journal = {Ann N Y Acad Sci},
  volume = {1000},
  pages = {205-221},
  note = {(2003,12)},
}

@article{Goh2020,
  author = {Goh, K. and Ng, C. and Lim, L. and Sheikh, U.},
  title = {Micro-expression recognition: an updated review of current trends, challenges and solutions},
  journal = {The Visual Computer},
  volume = {36},
  pages = {445-468},
  note = {(2020,3)},
  doi = {10.1007/s00371-018-1607-6},
}

@article{7851001,
  author = {Li, X. and Hong, X. and Moilanen, A. and Huang, X. and Pfister, T. and Zhao, G. and Pietikäinen, M.},
  year = {2018},
  title = {Towards Reading Hidden Emotions: A Comparative Study of Spontaneous Micro-Expression Spotting and Recognition Methods},
  journal = {IEEE Transactions On Affective Computing},
  volume = {9},
  pages = {563-577},
}

@article{Yildirim2023,
  author = {Yildirim, S. and Chimeumanu, M. and Rana, Z.},
  title = {The influence of micro-expressions on deception detection},
  journal = {Multimedia Tools And Applications},
  volume = {82},
  pages = {29115-29133},
  note = {(2023,8)},
  doi = {10.1007/s11042-023-14551-6},
}

@article{Frank2015,
  author = {Frank, M. and Svetieva, E. Microexpressions andDeception},
  year = {2015},
  title = {Understanding Facial Expressions In Communication: Cross-cultural And Multidisciplinary Perspectives},
  journal = {pp},
  pages = {227-242},
  doi = {10.1007/978-81-322-1934-7-11},
}

@article{Endres2009,
  author = {Endres, J. and Laidlaw, A.},
  title = {Micro-expression recognition training in medical students: a pilot study},
  journal = {BMC Medical Education},
  volume = {9},
  note = {47 (2009,7)},
  doi = {10.1186/1472-6920-9-47},
}

@article{WANG2024105091,
  author = {Wang, F. and Li, J. and Qi, C. and Wang, L. and Wang, P.},
  year = {2024},
  title = {JGULF: Joint global and unilateral local feature network for micro-expression recognition},
  journal = {Image And Vision Computing},
  volume = {147},
  pages = {105091},
}

@inproceedings{10.1145/3607829.3616446,
  author = {Guo, C. and Huang, H.},
  year = {2023},
  title = {GLEFFN: A Global-Local Event Feature Fusion Network for Micro-Expression Recognition},
  booktitle = {Proceedings Of The 3rd Workshop On Facial Micro-Expression},
  booksubtitle = {Advanced Techniques For Multi-Modal Facial Expression Analysis},
  pages = {17-24},
  doi = {10.1145/3607829.3616446},
}

@inproceedings{frank2009see,
  author = {Frank, M. and Herbasz, M. and Sinuk, K. and Keller, A. and Nolan, C.},
  year = {2009},
  title = {I see how you feel: Training laypeople and professionals to recognize fleeting emotions},
  booktitle = {The Annual Meeting Of The International Communication Association. Sheraton New York, New York City},
  pages = {1-35},
}

@article{9915437,
  author = {Li, Y. and Wei, J. and Liu, Y. and Kauttonen, J. and Zhao, G.},
  year = {2022},
  title = {Deep Learning for Micro-Expression Recognition: A Survey},
  journal = {IEEE Transactions On Affective Computing},
  volume = {13},
  pages = {2028-2046},
}

@inproceedings{10.1109/FG.2019.8756544,
  author = {Quang, N. and Chun, J. and Tokuyama, T.},
  year = {2019},
  title = {CapsuleNet for Micro-Expression Recognition},
  booktitle = {2019 14th IEEE International Conference On Automatic Face \& Gesture Recognition (FG 2019)},
  pages = {1-7},
  doi = {10.1109/FG.2019.8756544},
}

@inproceedings{8451376,
  author = {Li, Y. and Huang, X. and Zhao, G.},
  year = {2018},
  title = {Can Micro-Expression be Recognized Based on Single Apex Frame?},
  booktitle = {2018 25th IEEE International Conference On Image Processing (ICIP)},
  pages = {3094-3098},
}

@article{ACImage,
  author = {Verma, M. and Vipparthi, S. and Singh, G.},
  title = {Non-Linearities Improve OrigiNet based on Active Imaging for Micro Expression Recognition},
  journal = {2020,7)},
}

@inproceedings{Kumar2019ClassificationOF,
  author = {Kumar, A. and Theagarajan, R. and Peraza, O. and Bhanu, B.},
  year = {2019},
  title = {Classification of Facial Micro-expressions Using Motion Magnified Emotion Avatar Images},
  booktitle = {CVPR Workshops},
}

@article{10.1007/978-3-319-16865-4_34,
  author = {Wang, Y. and See, J. and Phan, R. and Oh, Y.},
  year = {2015},
  title = {LBP with Six Intersection Points: Reducing Redundant Information in LBP-TOP for Micro-expression Recognition},
  journal = {Computer Vision -- ACCV},
  volume = {2014.   \bibitem{inter}Huang, X., Wang, S., Zhao, G. \& Pietikäinen, M. Facial Micro-Expression Recognition Using Spatiotemporal Local Binary Pattern with Integral Projection. (2015,12)},
  pages = {525-537},
}

@article{HUANG2016564,
  author = {Huang, X. and Zhao, G. and Hong, X. and Zheng, W. and Pietikäinen, M.},
  year = {2016},
  title = {Spontaneous facial micro-expression analysis using Spatiotemporal Completed Local Quantized Patterns},
  journal = {Neurocomputing},
  volume = {175},
  pages = {564-578},
}

@inproceedings{Nguyen2023MicronBERTBF,
  author = {Nguyen, X. and Duong, C. and Li, X. and Gauch, S. and Seo, H. and Luu, K.},
  year = {2023},
  title = {Micron-BERT: BERT-Based Facial Micro-Expression Recognition},
  booktitle = {2023 IEEE/CVF Conference On Computer Vision And Pattern Recognition (CVPR)},
  pages = {1482-1492},
}

@inproceedings{10204934,
  author = {Fan, X. and Chen, X. and Jiang, M. and Shahid, A. and Yan, H.},
  year = {2023},
  title = {SelfME: Self-Supervised Motion Learning for Micro-Expression Recognition},
  booktitle = {2023 IEEE/CVF Conference On Computer Vision And Pattern Recognition (CVPR)},
  pages = {13834-13843},
}

@inproceedings{6553717,
  author = {Li, X. and Pfister, T. and Huang, X. and Zhao, G. and Pietikäinen, M.},
  year = {2013},
  title = {A Spontaneous Micro-expression Database: Inducement, collection and baseline},
  booktitle = {2013 10th IEEE International Conference And Workshops On Automatic Face And Gesture Recognition (FG)},
  pages = {1-6},
}

@article{Zeng2023,
  author = {Zeng, X. and Zhao, X. and Zhong, X. and Liu, G.},
  title = {A Survey of Micro-expression Recognition Methods Based on LBP, Optical Flow and Deep Learning},
  journal = {Neural Processing Letters},
  volume = {55},
  pages = {5995-6026},
  note = {(2023,10)},
  doi = {10.1007/s11063-022-11123-x},
}

@online{zhai2023featurerepresentationlearningadaptive,
  author = {Zhai, Z. and Zhao, J. and Long, C. and Xu, W. and He, S. and Zhao, H.},
  year = {2023},
  title = {Feature Representation Learning with Adaptive Displacement Generation and Transformer Fusion for Micro-Expression Recognition},
  url = {https://arxiv.org/abs/2304.04420},
}

@inproceedings{hu2018squeeze,
  author = {Hu, J. and Shen, L. and Sun, G.},
  year = {2018},
  title = {Squeeze-and-excitation networks},
  booktitle = {Proceedings Of The IEEE Conference On Computer Vision And Pattern Recognition},
  pages = {7132-7141},
}

@inproceedings{woo2018cbam,
  author = {Woo, S. and Park, J. and Lee, J. and Kweon, I.},
  year = {2018},
  title = {Cbam: Convolutional block attention module},
  booktitle = {Proceedings Of The European Conference On Computer Vision (ECCV)},
  pages = {3-19},
}

@article{of01,
  author = {Barron, J. and Fleet, D. and Beauchemin, S.},
  title = {Performance Of Optical Flow Techniques},
  journal = {International Journal Of Computer Vision},
  volume = {12},
  pages = {43-77},
  note = {(1994,2)},
}

@article{7286757,
  author = {Liu, Y. and Zhang, J. and Yan, W. and Wang, S. and Zhao, G. and Fu, X.},
  year = {2016},
  title = {A Main Directional Mean Optical Flow Feature for Spontaneous Micro-Expression Recognition},
  journal = {IEEE Transactions On Affective Computing},
  volume = {7},
  pages = {299-310},
}

@article{LIONG201882,
  author = {Liong, S. and See, J. and Wong, K. and Phan, R.},
  year = {2018},
  title = {Less is more: Micro-expression recognition from video using apex frame},
  journal = {Signal Processing: Image Communication},
  volume = {62},
  pages = {82-92},
}

@article{happy2017fuzzy,
  author = {Happy, S. and Routray, A.},
  year = {2017},
  title = {Fuzzy histogram of optical flow orientations for micro-expression recognition},
  journal = {IEEE Transactions On Affective Computing},
  volume = {10},
  pages = {394-406},
}

@inproceedings{kpcanet,
  author = {Feng, W. and Xu, M. and Chen, Y. and Wang, X. and Guo, J. and Dai, L. and Wang, N. and Zuo, X. and Li, X.},
  year = {2023},
  title = {Nonlinear Deep Subspace Network for Micro-expression Recognition},
  booktitle = {Proceedings Of The 3rd Workshop On Facial Micro-Expression},
  booksubtitle = {Advanced Techniques For Multi-Modal Facial Expression Analysis},
  pages = {1-8},
  doi = {10.1145/3607829.3616444},
}

@inproceedings{bilen2016dynamic,
  author = {Bilen, H. and Fernando, B. and Gavves, E. and Vedaldi, A. and Gould, S.},
  year = {2016},
  title = {Dynamic image networks for action recognition},
  booktitle = {Proceedings Of The IEEE Conference On Computer Vision And Pattern Recognition},
  pages = {3034-3042},
}

@inproceedings{Fernando_2015_CVPR,
  author = {Fernando, B. and Gavves, E. and Oramas, J. and Ghodrati, A. and Tuytelaars, T.},
  title = {Modeling Video Evolution for Action Recognition},
  booktitle = {Proceedings Of The IEEE Conference On Computer Vision And Pattern Recognition (CVPR)},
  address = {6},
  publisher = {2015},
}

@article{10981613,
  author = {Zhu, J. and Zong, Y. and Shi, J. and Lu, C. and Chang, H. and Zheng, W.},
  year = {2025},
  title = {Learning to Rank Onset-Occurring-Offset Representations for Micro-Expression Recognition},
  journal = {IEEE Transactions On Affective Computing},
  pages = {1-16},
}

@inproceedings{10446702,
  author = {Wang, L. and Huang, P. and Cai, W. and Liu, X.},
  year = {2024},
  title = {Micro-expression recognition by fusing action unit detection and Spatio-temporal features},
  booktitle = {ICASSP 2024 - 2024 IEEE International Conference On Acoustics, Speech And Signal Processing (ICASSP)},
  pages = {5595-5599},
}

@inproceedings{auasis,
  author = {Xie, H. and Lo, L. and Shuai, H. and Cheng, W.},
  year = {2020},
  title = {Au-assisted graph attention convolutional network for micro-expression recognition},
  booktitle = {Proceedings Of The 28th ACM International Conference On Multimedia},
  pages = {2871-2880},
}

@book{shao2025mol,
  author = {Shao, Z. and Cheng, Y. and Li, F. and Zhou, Y. and Lu, X. and Xie, Y. and Ma, L.},
  year = {2025},
  title = {Mol: Joint estimation of micro-expression, optical flow, and landmark via transformer-graph-style convolution},
  publisher = {IEEE Transactions On Pattern Analysis And Machine Intelligence},
}

@article{9194324,
  author = {Verma, M. and Vipparthi, S. and Singh, G.},
  year = {2021},
  title = {AffectiveNet: Affective-Motion Feature Learning for Microexpression Recognition},
  journal = {IEEE MultiMedia},
  volume = {28},
  pages = {17-27},
}

@article{soda4mer,
  author = {Bohao, Z. and Wang, X. and Wang, C. and He, G.},
  title = {Dynamic Stereotype Theory Induced Micro-expression Recognition with Oriented Deformation},
  journal = {2025,6)},
}

@inproceedings{9175230,
  author = {Lo, L. and Xie, H. and Shuai, H. and Cheng, W.},
  year = {2020},
  title = {MER-GCN: Micro-Expression Recognition Based on Relation Modeling with Graph Convolutional Networks},
  booktitle = {2020 IEEE Conference On Multimedia Information Processing And Retrieval (MIPR)},
  pages = {79-84},
}

@article{6130343,
  author = {Zhao, G. and Pietikainen, M.},
  year = {2007},
  title = {Dynamic Texture Recognition Using Local Binary Patterns with an Application to Facial Expressions},
  journal = {IEEE Transactions On Pattern Analysis And Machine Intelligence},
  volume = {29},
  pages = {915-928},
}

@article{Yan2014-fa,
  author = {Yan, W. and Li, X. and Wang, S. and Zhao, G. and Liu, Y. and Chen, Y. and Fu, X.},
  title = {CASME II: an improved spontaneous micro-expression database and the baseline evaluation},
  journal = {PLoS One},
  volume = {9},
  note = {e86041 (2014,1)},
}

@article{PAN2023106258,
  author = {Pan, H. and Xie, L. and Wang, Z.},
  year = {2023},
  title = {C3DBed: Facial micro-expression recognition with three-dimensional convolutional neural network embedding in transformer model},
  journal = {Engineering Applications Of Artificial Intelligence},
  volume = {123},
  pages = {106258},
}

@article{7492264,
  author = {Davison, A. and Lansley, C. and Costen, N. and Tan, K. and Yap, M.},
  year = {2018},
  title = {SAMM: A Spontaneous Micro-Facial Movement Dataset},
  journal = {IEEE Transactions On Affective Computing},
  volume = {9},
  pages = {116-129},
}

@article{ZHOU2022108275,
  author = {Zhou, L. and Mao, Q. and Huang, X. and Zhang, F. and Zhang, Z.},
  year = {2022},
  title = {Feature refinement: An expression-specific feature learning and fusion method for micro-expression recognition},
  journal = {Pattern Recognition},
  volume = {122},
  pages = {108275},
}

@online{zhang2019selfattentiongenerativeadversarialnetworks,
  author = {Zhang, H. and Goodfellow, I. and Metaxas, D. and Odena, A.},
  year = {2019},
  title = {Self-Attention Generative Adversarial Networks},
  url = {https://arxiv.org/abs/1805.08318},
}

@article{gan2019off,
  author = {Gan, Y. and Liong, S. and Yau, W. and Huang, Y. and Tan, L.},
  year = {2019},
  title = {OFF-ApexNet on micro-expression recognition system},
  journal = {Signal Processing: Image Communication},
  volume = {74},
  pages = {129-139},
}

@article{wang2020micro,
  author = {Wang, C. and Peng, M. and Bi, T. and Chen, T.},
  year = {2020},
  title = {Micro-attention for micro-expression recognition},
  journal = {Neurocomputing},
  volume = {410},
  pages = {354-362},
}

@inproceedings{wei2023cmnet,
  author = {Wei, M. and Jiang, X. and Zheng, W. and Zong, Y. and Lu, C. and Liu, J.},
  year = {2023},
  title = {Cmnet: contrastive magnification network for micro-expression recognition},
  booktitle = {Proceedings Of The AAAI Conference On Artificial Intelligence},
  publisher = {\textbf{37}},
  pages = {119-127},
}

@article{mellm,
  author = {Zhang, Z. and Zhao, S. and Liu, S. and Yin, S. and Mao, X. and Xu, T. and Chen, E.},
  title = {MELLM: Exploring LLM-Powered Micro-Expression Understanding Enhanced by Subtle Motion Perception},
  journal = {2025,5)},
}

@article{smola2004tutorial,
  author = {Smola, A. and Schölkopf, B.},
  year = {2004},
  title = {A tutorial on support vector regression},
  journal = {Statistics And Computing},
  volume = {14},
  pages = {199-222},
}

@article{ofvig,
  author = {Zhang, L. and Zhang, Y. and Sun, X. and Tang, W. and Wang, X. and Li, Z.},
  year = {2025},
  title = {Micro-expression recognition based on direct learning of graph structure},
  journal = {Neurocomputing},
  volume = {619},
  pages = {129135},
}

@article{mersupcon,
  author = {Zhi, R. and Hu, J. and Wan, F.},
  year = {2022},
  title = {Micro-expression recognition with supervised contrastive learning},
  journal = {Pattern Recognition Letters},
  volume = {163},
  pages = {25-31},
}

@article{9363624,
  author = {Gupta, P.},
  year = {2023},
  title = {MERASTC: Micro-Expression Recognition Using Effective Feature Encodings and 2D Convolutional Neural Network},
  journal = {IEEE Transactions On Affective Computing},
  volume = {14},
  pages = {1431-1441},
}

@inproceedings{liu2020offset,
  author = {Liu, N. and Liu, X. and Zhang, Z. and Xu, X. and Chen, T.},
  year = {2020},
  title = {Offset or onset frame: A multi-stream convolutional neural network with capsulenet module for micro-expression recognition},
  booktitle = {2020 5th International Conference On Intelligent Informatics And Biomedical Sciences (ICIIBMS)},
  pages = {236-240},
}

@article{8844867,
  author = {Verma, M. and Vipparthi, S. and Singh, G. and Murala, S.},
  year = {2020},
  title = {LEARNet: Dynamic Imaging Network for Micro Expression Recognition},
  journal = {IEEE Transactions On Image Processing},
  volume = {29},
  pages = {1618-1627},
}

@article{e25030460,
  author = {Zhou, H. and Huang, S. and Li, J. and Wang, S.},
  year = {2023},
  title = {Dual-ATME: Dual-Branch Attention Network for Micro-Expression Recognition},
  journal = {Entropy},
  volume = {25},
  url = {https://www.mdpi.com/1099-4300/25/3/460},
}

@article{HE2025128372,
  author = {He, J. and Xiao, Y. and Zhang, H. and Cai, J. and Cai, L. and Liu, R.},
  year = {2025},
  title = {Micro\_NesT: multi-scale attention enhanced micro-expression recognition framework},
  journal = {Expert Systems With Applications},
  volume = {290},
  pages = {128372},
}

@inproceedings{lo2020mer,
  author = {Lo, L. and Xie, H. and Shuai, H. and Cheng, W.},
  year = {2020},
  title = {MER-GCN: Micro-expression recognition based on relation modeling with graph convolutional networks},
  booktitle = {2020 IEEE Conference On Multimedia Information Processing And Retrieval (MIPR)},
  pages = {79-84},
}

@article{li2020joint,
  author = {Li, Y. and Huang, X. and Zhao, G.},
  year = {2020},
  title = {Joint local and global information learning with single apex frame detection for micro-expression recognition},
  journal = {IEEE Transactions On Image Processing},
  volume = {30},
  pages = {249-263},
}

@article{NIE202113,
  author = {Nie, X. and Takalkar, M. and Duan, M. and Zhang, H. and Xu, M.},
  year = {2021},
  title = {GEME: Dual-stream multi-task GEnder-based micro-expression recognition},
  journal = {Neurocomputing},
  volume = {427},
  pages = {13-28},
}

@inproceedings{9747232,
  author = {Wei, M. and Zheng, W. and Zong, Y. and Jiang, X. and Lu, C. and Liu, J.},
  year = {2022},
  title = {A Novel Micro-Expression Recognition Approach Using Attention-Based Magnification-Adaptive Networks},
  booktitle = {ICASSP 2022 - 2022 IEEE International Conference On Acoustics, Speech And Signal Processing (ICASSP)},
  pages = {2420-2424},
}

\end{document}